\documentclass[]{fairmeta}
\usepackage{xcolor}
\usepackage{longtable}
\usepackage{listings}
\usepackage[numbers]{natbib}
\definecolor{codegreen}{rgb}{0,0.6,0}
\definecolor{codegray}{rgb}{0.5,0.5,0.5}
\definecolor{codepurple}{rgb}{0.58,0,0.82}
\definecolor{backcolour}{rgb}{0.95,0.95,0.92}
\definecolor{promptcolor}{HTML}{D1D0F2}
\definecolor{promptcolorheader}{HTML}{bdbcec}
\newcommand{\promptbox}[2]{
\begin{tcolorbox}[
top=0.3em,bottom=0.3em,left=0.5em,right=0.5em,
toptitle=0.3em,bottomtitle=0.2em,boxsep=0pt,
colframe=promptcolorheader,colback=promptcolor!50,boxrule=0.5pt,
title={\footnotesize \fontfamily{zi4}\selectfont #1}
]
\footnotesize
{\fontfamily{zi4}\selectfont #2}
\end{tcolorbox}
}
\lstdefinestyle{mystyle}{
    backgroundcolor=\color{backcolour},   
    commentstyle=\color{codegreen},
    keywordstyle=\color{magenta},
    numberstyle=\tiny\color{codegray},
    stringstyle=\color{codepurple},
    basicstyle=\ttfamily\footnotesize,
    breakatwhitespace=false,         
    breaklines=true,                 
    captionpos=b,                    
    keepspaces=true,                 
    numbers=left,                    
    numbersep=5pt,                  
    showspaces=false,                
    showstringspaces=false,
    showtabs=false,                  
    tabsize=2
}

\title{CausalVQA: A Physically Grounded Causal Reasoning Benchmark for Video Models}

\author[1]{Aaron Foss}
\author[1]{Chloe Evans}
\author[1]{Sasha Mitts}
\author[1]{Koustuv Sinha}
\author[1]{Ammar Rizvi}
\author[1]{Justine T. Kao}

\affiliation[1]{FAIR at Meta}

\abstract{We introduce CausalVQA, a benchmark dataset for video question answering (VQA) composed of question-answer pairs that probe models' understanding of causality in the physical world. Existing VQA benchmarks either tend to focus on surface perceptual understanding of real-world videos, or on narrow physical reasoning questions created using simulation environments. CausalVQA fills an important gap by presenting challenging questions that are grounded in real-world scenarios, while focusing on models' ability to predict the likely outcomes of different actions and events through five question types -- \textit{counterfactual}, \textit{hypothetical}, \textit{anticipation}, \textit{planning} and \textit{descriptive}. We designed quality control mechanisms that prevent models from exploiting trivial shortcuts, requiring models to base their answers on deep visual understanding instead of linguistic cues. We find that current frontier multimodal models fall substantially below human performance on the benchmark, especially on anticipation and hypothetical questions. This highlights a challenge for current systems to leverage spatial-temporal reasoning, understanding of physical principles, and comprehension of possible alternatives to make accurate predictions in real-world settings.
}

\date{\today}
\correspondence{Aaron Foss at \email{afoss@meta.com}}

\metadata[Code]{\url{https://github.com/facebookresearch/CausalVQA}}
\metadata[Blogpost]{\url{https://ai.meta.com/blog/v-jepa-2-world-model-benchmarks}}

\begin{document}

\maketitle

\section{Introduction}
\label{section:intro}
Humans possess an intuitive understanding of the physical world. We use it to make predictions and plan, imagining and simulating what will likely happen (e.g. if a vase is placed close to a table’s edge, it might fall and break when someone bumps into the table) without needing to go through costly iterations of trial and error. As AI moves to non-stationary form factors and wearable devices that follow us as we navigate physical objects and space, we will require models that have a robust understanding of the physical world; we will want models capable of proactively providing suggestions, anticipating outcomes, and generally helping users accomplish their real-world goals.

With advancements in recent years, large multimodal models have achieved remarkable performance on various visual capabilities \citep{geminiteam2024gemini15unlockingmultimodal, zhang2024mmllmsrecentadvancesmultimodal, gpt4v}. Many video-based tasks have been developed to assess models from a range of perspectives, such as action recognition \citep{kay2017kineticshumanactionvideo}, video summarization \citep{han2025shot2storynewbenchmarkcomprehensive}, social reasoning \citep{0001CLTM19, NEURIPS2023_7a65606f}, spatial reasoning \citep{yu2019activitynetqadatasetunderstandingcomplex}, episodic memory \citep{OpenEQA2023}, and longform video understanding \citep{mangalam2023egoschemadiagnosticbenchmarklongform, chandrasegaran2024hourvideo1hourvideolanguageunderstanding, rawal2024cinepilelongvideoquestion}. These benchmarks tend to be based on diverse real videos, either taken ``in the wild'' \citep{grauman2022ego4dworld3000hours}, selected from existing film clips \citep{rawal2024cinepilelongvideoquestion, lei2019tvqalocalizedcompositionalvideo}, or collected following a set of scripts and instructions \citep{pătrăucean2023perceptiontestdiagnosticbenchmark}. The majority of these real-video based benchmarks focus on perceptual and descriptive aspects of visual tasks, with very few focusing on causality in robust and systematic ways.

An alternative approach has resulted in a style of benchmarks that focus specifically on evaluating models' causal understanding from visual input, built using fully-controlled synthetic environments with simulated videos that contain a relatively small set of objects and actions \citep{yi2020clevrercollisioneventsvideo, chen2022comphycompositionalphysicalreasoning, zheng2024contphy, patel2022crippvqacounterfactualreasoningimplicit}. 
While these benchmarks can carefully remove biases and isolate causal reasoning, they (by design) lack the diversity and complexity that make performing physical causal reasoning in the real world challenging.

In order to evaluate multimodal models' out-of-the-box causal reasoning abilities in real-world settings, we introduce CausalVQA, a video multiple-choice question (MCQ) dataset built from egocentric videos selected from EgoExo4D \citep{grauman2024egoexo4dunderstandingskilledhuman}. We designed a hybrid human-and-model pipeline for creating and curating the question and answer pairs, which helped maintain a high bar for diversity and real-world relevance while preventing models from relying on superficial visual and linguistic cues to perform well on the task. The data creation pipeline resulted in $1,786$ items across five question categories: \textit{counterfactual}, \textit{hypothetical}, \textit{anticipation}, \textit{planning} and \textit{descriptive}.

We evaluated the zero-shot capabilities of state-of-the-art multimodal models on CausalVQA, including closed models such as GPT-4o \citep{gpt4o} and Gemini 2.5 Flash \cite{gemini} and frontier open models such as Qwen2.5VL \citep{bai2025qwen25vltechnicalreport} and PerceptionLM \citep{cho2025PerceptionLM}. We also conducted a particularly comprehensive human evaluation by asking $273$ naive annotators (untrained for the task) to answer the same set of questions. We used these results to assign empirical difficulty levels to each item in the dataset. Humans achieved $84.78\%$ accuracy on the benchmark ($95.05\%$, $77.32\%$, and $60.05\%$ on \textit{easy}, \textit{medium}, and \textit{hard} subsets, respectively), while the highest-performing model we evaluated (Gemini 2.5 Flash) scored $61.66\%$ overall. These results show that there is still a meaningful gap between model and human performance on visual causal reasoning in real-world videos.

\begin{table*}[t]
    \centering
    \begin{NiceTabular}{lm{35em}}
    \CodeBefore
    \Body
    \toprule
   Question Type & Description \\
   \midrule
   Counterfactual & Questions pertaining to what could have happened if a something had changed or had occurred differently in the clip.\\
     \hline
   Hypothetical & Questions pertaining to what else could potentially happen in the same scenario as presented in the video. Compared to counterfactual questions, hypothetical questions tend to be only loosely connected to specific events in the scenario.\\
   \hline
   Anticipation & Questions regarding what is likely to happen next in the video, either based on laws of physics (e.g. the trajectory of a ball) or common patterns in human behavior (e.g. reaching to catch a ball). \\
     \hline
     Planning & Questions about the steps needed in order to achieve an objective, or the reasons why someone in the clip performs an action (in order to advance towards their goal).\\
   \hline
   Descriptive & Questions about the facts presented in the video (e.g. actions performed, order of actions, salient objects).\\
    
    \bottomrule
    \end{NiceTabular}
    \caption{Definitions of each of the five question types annotators were asked to create for each clip. With the exception of \textit{descriptive} questions, all other question types involve some form of causal reasoning, and are collectively referred to as \textit{Reasoning} questions. A more detailed version of the question definitions was included in guidelines shared with annotators (see Appendix \ref{question_guidelines}).}
    \label{table:question_types}
\end{table*}

\section{The CausalVQA Dataset}
\label{section:methods}

This section describes the design principles and creation process for CausalVQA. We outline the steps taken to ensure scene diversity, question quality, and alignment with human responses, concluding with a discussion of the resulting benchmark's composition.

\subsection{Design Objectives}
The design choices we made when constructing CausalVQA were based on the following desiderata:
\begin{itemize}

\item \textbf{Real-world relevance}. Focus on scenarios/questions closely tied to real-world use cases and applications.

\item \textbf{Physical phenomena}. Emphasize causal understanding in the physical world, rather than descriptions of events or reasoning about mental states such as goals and beliefs.

\item \textbf{Reduce susceptibility to shortcuts}. Require true comprehension of the video input and avoid questions that can be answered by relying only on world knowledge encoded in a text-only, ``blind'' Large Language Model (LLM).

\item \textbf{Easy for humans}. Adhere closely to lay-people's intuitive understanding of the physical world and the relationships between objects and activities.

\item \textbf{Ease of evaluation}. Ensure relatively easy and inexpensive evaluation, with clear metrics and a manageable set of items for assessing zero-shot performance.
\end{itemize}

Based on these design objectives, we developed a process for creating and curating a high-quality set of videos and question-answer pairs targeting causal reasoning in the complex physical world. We believe the resulting benchmark fills an important gap in existing literature, enabling focused measurement of models' understanding of real-world dynamics and cause-and-effect that are as intuitive as they are critical to humans. We describe the full generation and curation process in the following section.

\begin{figure}[!h]
    \centering
    
    \label{fig:enter-label}
    \includegraphics[width=1\linewidth]{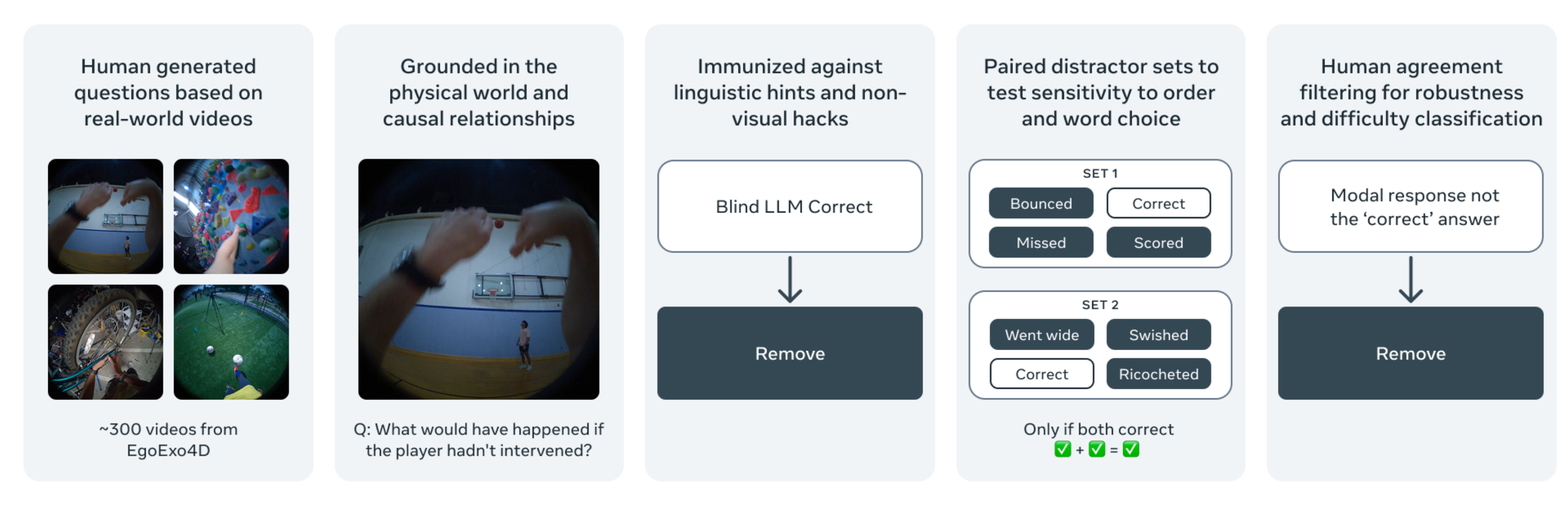}
    \caption{Process of generating and curating question-answer pairs for CausalVQA. Multiple steps were incorporated in order to ensure diversity and visual groundedness as well as reduce susceptibility to shortcuts.}
\end{figure}

\subsection{Generation Process}
We developed a multi-stage pipeline that leverages both human expertise and automated tools. The process was designed to produce high-quality, realistic, and diverse questions, while reducing the potential for models to select correct answers through spurious correlations.

\subsubsection{Video Selection}

We first selected $298$ egocentric videos from the EgoExo4D dataset as the source material, which are readily available under the EgoExo4D License. These videos feature a diverse range of familiar real-world scenarios and activities with rich interactions between the camera-wearer and the physical world around them, making them an ideal choice to satisfy \textbf{real-world relevance}.

We addressed \textbf{physical phenomena} by choosing videos of various physical activities, such as sports and cooking. We sought scenarios that are goal-directed (e.g., climbing an indoor rock wall or practicing a basketball shot), self-contained, and easy to understand; we avoided scenarios that contain relatively few interactions between the camera-wearer and other objects (e.g. solo dance or music practice), to provide a richer substrate for cause-and-effect question creation. Manual filtering on EgoExo4D metadata resulted in roughly 12 hours worth of selected videos.

\subsubsection{Initial Question Generation}

Drawing from literature on causal reasoning \citep{hagmayer2007causal, kulakova2013processing} and in line with taxonomies employed in related benchmarks \citep{yi2020clevrercollisioneventsvideo, li2022representationreasoningevidencecommonsense}, we outlined five question types for our benchmark: \textit{counterfactual}, \textit{hypothetical}, \textit{anticipation}, \textit{planning}, and \textit{descriptive} (see Table \ref{table:question_types}). Our main focus was on probing models' performance on the first four question types, but we included \textit{descriptive} questions to enable comparison between models' performance on questions that require in-depth causal reasoning versus more surface-level visual perception. As a shorthand, we will refer to the question types excluding \textit{descriptive} ones as \textit{Reasoning} questions.

We recruited a team of $16$ annotators through an outside vendor. Prior to the task, annotators were given a set of definitions and question-type examples. They were also coached to ensure their questions were ``visually grounded'' (i.e., specific visual information presented in the video is required to answer the question). Since our goal was to capture a natural and relevant question set, annotators had ample freedom to use creativity and intuition when generating questions (see Appendix \ref{question_guidelines} for guidelines). This is in contrast to other benchmarks that employed templates for the question generation process \citep{pătrăucean2023perceptiontestdiagnosticbenchmark, chandrasegaran2024hourvideo1hourvideolanguageunderstanding}.

Each video was seen by three separate annotators, and each annotator generated $3$-$5$ question-and-answer pairs across the question types. For each question, annotators supplied a start and end timestamp and were instructed to only include the exact video segment that is required to answer the question. This step resulted in $3,764$ questions.

\subsubsection{Distractor Generation}

We formulate the task as a set of multiple-choice questions (MCQs), which helps satisfy \textbf{ease of evaluation} and avoids the complexities of evaluating the correctness of free-form text. In order to convert the question-answer pairs generated from annotators in the first step into MCQs, we generated incorrect but plausible options -- referred to as ``distractors'' -- using an automated pipeline. 

For each video segment identified and annotated in the previous step, we asked a vision language model (VLM) to describe the activity in the clip. We separately prompted the VLM to describe the activity occurring up to three seconds after the clip's end (referred to as post-clip), which was then used in subsequent steps to refine anticipation questions. We then prompted the VLM to create a set of initial `seed' distractors by providing it with the clip, text descriptions of the clip and post-clip, as well as instructions to generate wrong answers that reference elements that are visible in the clips.

\subsubsection{Language Refinement}

Once each video segment and human-generated question/answer has been augmented with video captions and seed distractor answers, we used an LLM (Llama 3.1-70B-Instruct \citep{grattafiori2024llama3herdmodels}) to refine and normalize the language, in order to \textbf{reduce susceptibility to shortcuts}. We prompted the LLM to paraphrase the annotator-provided question and answer in consistent and grammatical English. We also asked it to remove words that may contain hints about the answer and replace them with more neutral phrasings. Throughout the process, we performed automatic checks with the LLM to ensure the question's meaning remained intact. See Appendix \ref{app:language_refinement} for additional details on the LLM prompts.

We also used the LLM to enhance the set of distractors by prompting the model with both the VLM-generated seed distractors and descriptions of the videos. This step resulted in a set of more diverse and higher-quality distractors as the VLM tends towards a very narrow range of alternatives (see Table~\ref{fig:distractor-refinement}).

\begin{table}[h]
  \centering
  \small 
  \begin{tabular}{p{0.5\textwidth}p{0.5\textwidth}}
    \hline
    \textbf{VLM Seed Distractors} & \textbf{Refined Distractors} \\
    \hline
    \begin{minipage}[t]{\linewidth}
      \vspace{1mm}
      1. The person will turn to the left next \\
      2. The person will turn to the left and then to the right \\
      3. The person will turn to the right and then to the left \\
      4. The person will turn to the left and then to the left again
      \vspace{1mm}
    \end{minipage} &
    \begin{minipage}[t]{\linewidth}
      \vspace{1mm}
      1. The person will advance to strike the ball \\ 
      2. The person will retreat to gain distance \\
      3. The person will pivot to the left to evade a challenger\\
      4. The person will leap up to meet the ball
      \vspace{1mm}
    \end{minipage} \\
    \hline
  \end{tabular}
  \caption{An example of distractors generated directly by a VLM, compared to distractors refined in a further step using Llama 3.1-70B-Instruct. The refinement step resulted in more diverse items, as well as the replacement of the original Option 3, which could have been an alternative correct answer (i.e., `the person will turn to the right').}
  \label{fig:distractor-refinement}
\end{table}

\subsubsection{Language Perturbation and Reordering}
\label{sec:perturb}
Based on the observations that model performance on MCQs can be sensitive to the set of distractors provided \citep{alhazmi2024distractorgenerationmultiplechoicetasks} as well as the order in which answer options are presented \citep{pezeshkpour2023largelanguagemodelssensitivity}, we incorporated a perturbation mechanism into the benchmark design that penalizes models for over-indexing on surface characteristics of the distractors, thus more precisely measuring models' ability to understand and reason about the videos themselves. 

We perturbed the MCQs by using the LLM to generate paraphrases of each distractor, followed by reshuffling the order of the correct answer and the paraphrased distractors. This resulted in two variations of the same set of questions, which we refer to as $MCQ$ and $MCQ'$ (see Table~\ref{fig:mcq-sets}). Based on these alternative answer sets, we devised two metrics for model performance -- \textit{unpaired}, which counts the model as correct if it answers a $MCQ$ correctly, and \textit{paired}, which counts the model as correct only if it answers both $MCQ$ and $MCQ'$ correctly. The two versions of questions and \textit{unpaired} vs \textit{paired} accuracy allow benchmark users to quantify models' ability to robustly answer the questions. Under certain assumptions, doubling the number of items in this way while computing paired metrics reduces the theoretical variance of a model's response to less than $1/4$ (see Appendix \ref{app:pairing}), substantially increasing the benchmark's statistical power even with a relatively small set of items. Results reported in the paper are \textit{paired} unless otherwise noted.

\begin{table}[h]
  \centering
  \small 
  \begin{tabular}{p{0.5\textwidth}p{0.5\textwidth}}
    \hline
    \textbf{MCQ} & \textbf{MCQ'} \\
    \hline
    \begin{minipage}[t]{\linewidth}
      \vspace{1mm}
      A. The person will move backward to create space \\
      B. The person will turn to the right next \\
      C. The person will move forward to kick the ball \\
      D. The person will turn to the left to avoid an opponent \\
      E. The person will jump up to head the ball
      \vspace{1mm}
    \end{minipage} &
    \begin{minipage}[t]{\linewidth}
      \vspace{1mm}
      A. The person will turn to the right next \\
      B. The person will leap up to meet the ball \\
      C. The person will retreat to gain distance \\
      D. The person will advance to strike the ball \\
      E. The person will pivot to the left to evade a challenger
      \vspace{1mm}
    \end{minipage} \\
    \hline
  \end{tabular}
  \caption{Example of the original Multiple Choice Question (\texttt{MCQ}) option set and its perturbed and re-ordered version (\texttt{MCQ'}). Each option in \texttt{MCQ'} is a paraphrase of an option in \texttt{MCQ}, with the exception of the actual correct answer, which is identical in both versions. There is no statistically significant difference between human performance on the set of \texttt{MCQ} compared to \texttt{MCQ'}, suggesting that the perturbation and re-ordering mechanism we used did not inadvertantly change the difficulty of the original questions.}
  \label{fig:mcq-sets}
\end{table}

\subsubsection{Ensuring Visual Groundedness}

A critical design objective of our benchmark is to focus on questions that require models to understand the visual information in order to arrive at the correct answer. To that end, we included a filtering step to remove all items in which a text-only LLM (again Llama 3.1-70B-Instruct) \cite{grattafiori2024llama3herdmodels} can answer correctly without ``seeing'' any of the videos. We ran the LLM on both versions of each question, and removed the items if the LLM was able to answer both versions correctly without any access to the video segment. 

While this is an aggressive filtering step (approximately $39\%$ of our questions were removed), it substantially reduces the number of items that can be trivially solved with the world knowledge encoded in a typical model's language backbone, allowing the benchmark to focus more precisely on questions that require visual understanding.

\subsubsection{Human Quality Assurance}

We conducted a further round of human quality assurance with our vendor in order to detect and repair any errors introduced in each previous step of the process. Each version of each question was reviewed by two annotators, who checked for the presence of six types of errors -- (a) misaligned timestamp, (b) question does not require video to answer, (c) answer to question is incorrect, (d) incorrect question type, (e) question and answer do not make sense, and (f) question has multiple correct answers. We removed questions where reviewers flagged errors (a)-(c), as they tend to have more fundamental issues, and repaired questions where at most one reviewer flagged errors (d)-(f) through a combination of programmatic checks and manual correction. This final step removed $2,306$ items ($53\%$) and maintained our quality bar.

\subsubsection{Human-Derived Empirical Difficulty Levels}

Our final step was to assign difficulty levels to each item remaining in the benchmark based on empirical human performance, allowing us to better understand how models perform on questions that are \textbf{easy for humans}, versus ones people would also find challenging or ambiguous.

This step required a large-scale human baseline evaluation with $273$ people who were not previously exposed to the task. Each version of the question was presented to $15$ annotators, who independently selected the answer they deemed most correct out of the $5$ options. We verified that the modal response for each question matched the ``ground truth'' answer, and removed $~14\%$ of items where this was not the case, likely due to errors in the question formulation that were not caught in the previous quality assurance step. Finally, we verified that any difference between the two versions of each question was not statistically significant.  

We then computed an agreement rate for each question by as the percentage of people whose response agreed with the modal/correct response. We labeled an item as \textit{easy} if agreement was $>86.67\%$, \textit{medium} if agreement was between $70\%$ and $86.67\%$, and hard if agreement was $36.67\%$ to $70\%$. Table \ref{tab:revised} contains examples of the resulting difficulty assignment, and Table \ref{table:dist} shows the benchmark's  difficulty breakdown.

\begin{table}[!htb]
\centering
\small
\begin{tabular}{p{0.1\linewidth}p{0.25\linewidth}p{0.25\linewidth}p{0.25\linewidth}}
\hline
\multicolumn{1}{l}{\textbf{Difficulty}} & \multicolumn{1}{c}{Easy} & \multicolumn{1}{c}{Medium} & \multicolumn{1}{c}{Hard} \\
\hline
\multicolumn{1}{l}{\textbf{Category}} & \multicolumn{1}{c}{Counterfactual} & \multicolumn{1}{c}{Anticipation} & \multicolumn{1}{c}{Hypothetical} \\
\hline
\textbf{Frame} & 
\includegraphics[width=\linewidth]{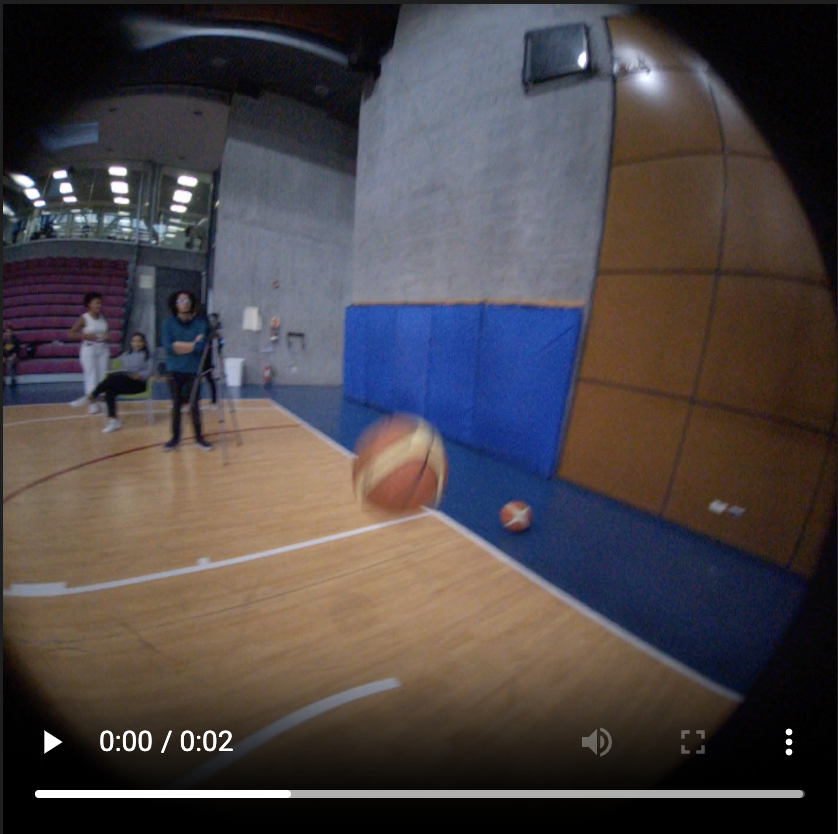} & 
\includegraphics[width=\linewidth]{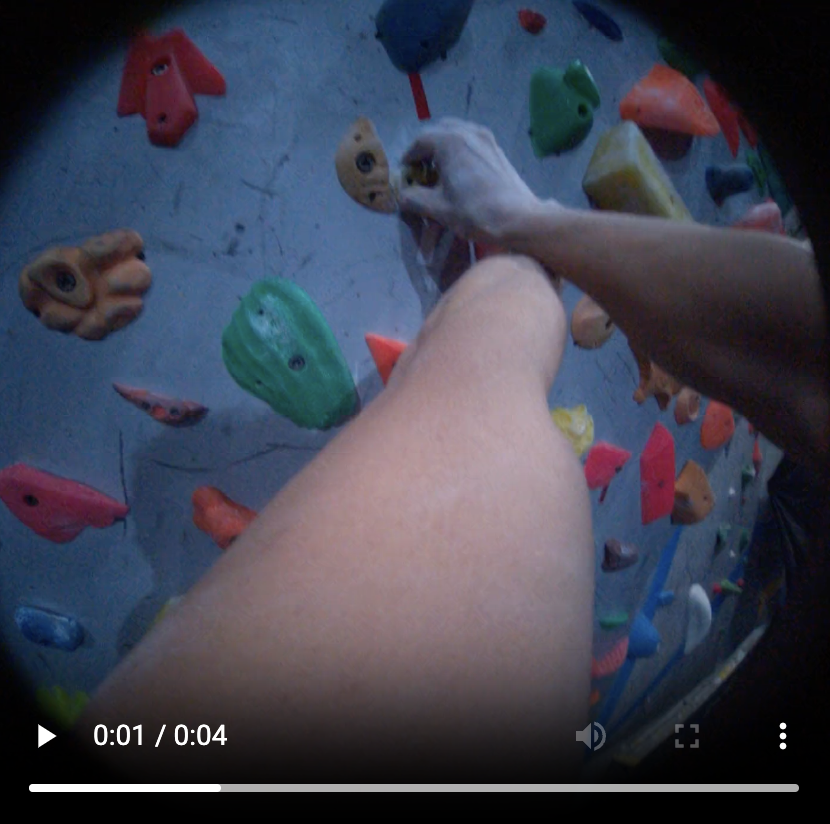} & 
\includegraphics[width=\linewidth]{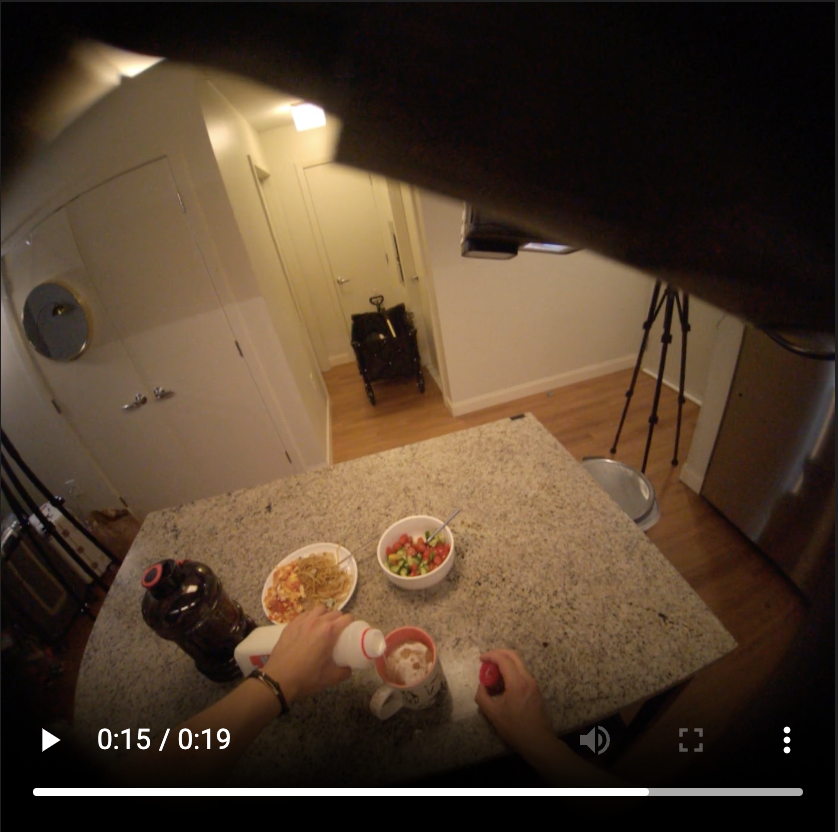} \\
\hline
\textbf{Question / Answer} & 
\begin{minipage}[t]{\linewidth}
\vspace{1mm}
If the object the player is using gets away, is there another object available? \\
A. The player can use the ball that is currently being used by another player on the court. \\
B. There is a ball rack with multiple balls located near the wall, but it's out of the player's reach. \\
C. There is another ball available on the balcony above the court. \\
D. There is a spare ball located near the entrance of the gymnasium. \\
\textit{E. There is another ball available on the blue floor, closer to the wall.}
\vspace{1mm}
\end{minipage} & \begin{minipage}[t]{\linewidth}
\vspace{1mm}
In which direction will the person move next? \\
A. The person will move in a zigzag pattern to the right next. \\ \textit{B. The person will head upwards and to the left next.} \\ C. The person will shift their weight to the left and move horizontally next. \\ D. The person will move in a circular motion to the left next. \\ E. The person will move diagonally to the right and downwards next \\
\vspace{1mm}
\end{minipage} & \begin{minipage}[t]{\linewidth}
\vspace{1mm}
Is there enough of the beverage ingredient left for another serving? \\
A. There is enough cereal for another serving. \\
B. There is no beverage ingredient left. \\
\textit{C. There is enough milk for another tea.} \\
D. There is just enough water for one more cup. \\
E. It is unclear if there is enough beverage ingredient.
\vspace{1mm}
\end{minipage} \\
\hline
\textbf{Notes} & Requires recognizing another object within the field of vision. & Requires noticing the movement of the hands and field of view. & Requires identifying what is being served, which beverage is being referred to, and the amount consumed and remaining. \\
\hline
\end{tabular}
\caption{Example questions across question types and difficulty levels. The correct answer is in \textit{italics}.}
\label{tab:revised}
\end{table}

\begin{table}[!htb]
    \centering

    \label{tab:main_table}
    
    \begin{subtable}[t]{0.3\textwidth}
        \centering
        \label{tab:subtable_1}
        \begin{tabular}{lr}
            \toprule
           Question Type & Count\\
           \midrule
           Descriptive & $288$ \\
           Anticipation & $208$ \\
           Planning & $141$ \\
           Counterfactual & $89$ \\   
           Hypothetical & $67$ \\
            \bottomrule
        \end{tabular}
    \end{subtable}%
    \begin{subtable}[t]{0.3\textwidth}
        \centering
        \label{tab:subtable_2}
        \begin{tabular}{lr}
            \toprule
            Difficulty & Count \\
            \midrule
            Easy & $454$ \\
            Medium & $215$ \\
            Hard & $124$ \\
            \bottomrule
        \end{tabular}
    \end{subtable}
    \begin{subtable}[t]{0.3\textwidth}
        \centering
        \label{tab:subtable_3}
        \begin{tabular}{lr}
            \toprule
           Scenario & Count\\
           \midrule
           Basketball & $211$ \\
           Rock Climbing & $181$ \\
           Cooking & $174$ \\
           Bike Repair & $111$ \\   
           Soccer & $86$ \\
           Health & $28$ \\
            \bottomrule
        \end{tabular}
    \end{subtable}%

    \caption{Benchmark composition by question type, difficulty level, and scenario. \textit{Reasoning} (i.e. non-\textit{descriptive}) questions make up $63.7\%$ of the benchmark, and $57.2\%$ of questions are considered \textit{Easy} for humans based on the high agreement rate among $15$ naive annotators. The main scenarios represented in the benchmark are related to sports ($60.4\%$), followed by other scenarios with complex and causally connected sequences of actions such as cooking and bike repair.}
    \label{table:dist}
\end{table}

\subsection{Dataset Description}

\subsubsection{Benchmark Composition}

\begin{figure}[!h]
  \begin{subfigure}[b]{0.45\textwidth}
    \includegraphics[width=\textwidth]{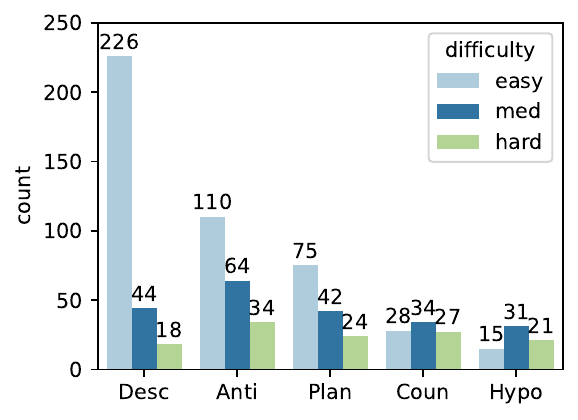}
    \caption{Number of question pairs for each question category and difficulty level.}
  \end{subfigure}
  \hspace{0.05\textwidth}
  \begin{subfigure}[b]{0.45\textwidth}
    \includegraphics[width=\textwidth]{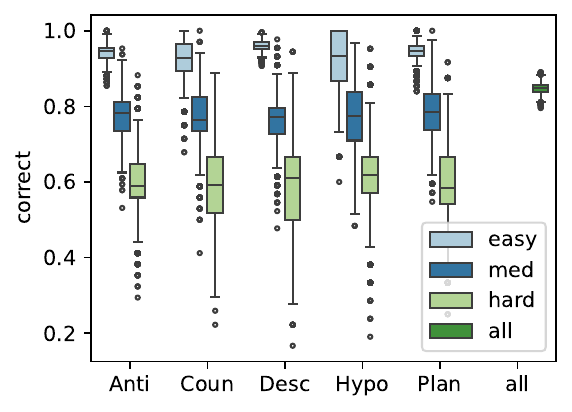}
    \caption{Bootstrapped human performance for each question type and difficulty level.}
  \end{subfigure}
  \caption{Breakdowns of question pairs by question category and difficulty level.}
  \label{fig:counts}
\end{figure}

\begin{figure}[!h]
  \begin{subfigure}[b]{0.45\textwidth}
    \includegraphics[width=\textwidth]{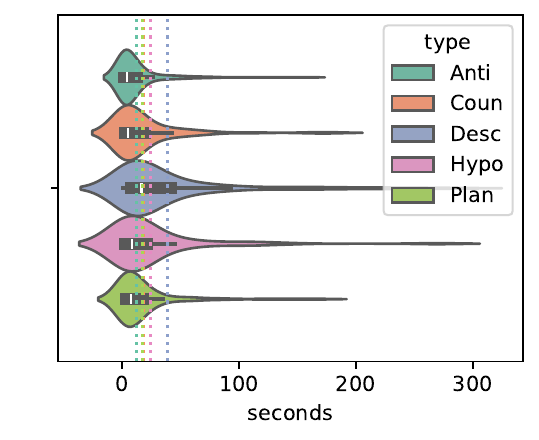}
    \caption{Clip duration distribution by question category. Dotted vertical lines indicate means.}
  \end{subfigure}
  \hspace{0.05\textwidth}
  \begin{subfigure}[b]{0.45\textwidth}
    \includegraphics[width=\textwidth]{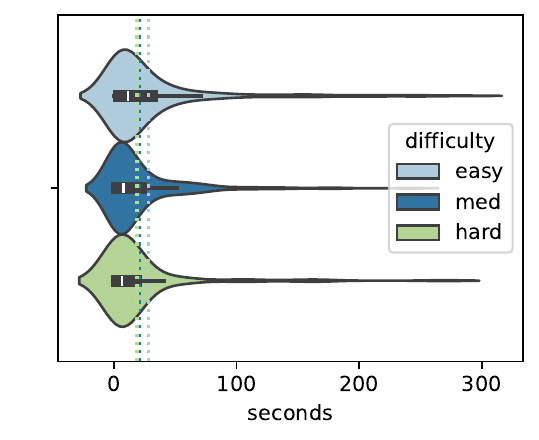}
    \caption{Clip duration distribution by difficulty level. Dotted vertical lines indicate means.}
  \end{subfigure}
  \caption{Videos in each question type and difficulty level have generally similar and overlapping duration distributions.}
  \label{fig:durations}
\end{figure}

The final benchmark contains 1,586 items (793 paired questions) and 779 video segments. We provide a breakdown of the benchmark's composition in Table \ref{table:dist} and Figures \ref{fig:counts} and \ref{fig:durations}. These statistics illustrate the variety and distribution of question pairs across several meaningful dimensions.  While a more balanced dataset may be desired, our yields were dictated by the generating process -- easy hypothetical questions, for whatever reason, are hard to write in ways that are robust to the shortcuts we sought to avoid. However, our pairing process (see \ref{sec:perturb}) partially ameliorates this, giving the benchmark more statistical power than the length alone would indicate.

Our average clip duration is 24.8 seconds, with a maximum of 288.9 seconds. Additionally, as seen in Figure \ref{fig:durations}(b), there is considerable overlap in question duration by question category and difficulty. This suggests that the benchmark's difficulty stems from the complexity of the queries and content, rather than requirements on memory and ability to process longer videos. This distinguishes CausalVQA from other benchmarks such as EgoSchema \citep{mangalam2023egoschemadiagnosticbenchmarklongform} and HourVideo \citep{chandrasegaran2024hourvideo1hourvideolanguageunderstanding}, where the length of the video is a primary source of challenge for models.

\subsubsection{Debug Dataset}
In addition to the main benchmark, we set aside a ``debug'' dataset consisting of 100 questions (20 from each question category, all from the easy difficulty level). This subset is designed to facilitate model parameter selection and is not included in the final benchmark statistics described above, or the evaluation results reported in the paper. By providing this separate dataset, we aim to support researchers in their efforts to use the CausalVQA benchmark without inadvertently over-fitting on the main test set.

\subsubsection{Qualitative Assessment}
To ensure that the benchmark design and content align well with key intuitive elements of physical reasoning, we conducted a qualitative user study with $12$ participants who were experienced with using AI products, but were otherwise not familiar with existing benchmarks in the space. Participants were first asked to describe what they believe are important aspects of reasoning in the physical world. They were then shown a random selection of items from CausalVQA and asked whether the questions aligned with their definition of physical reasoning. Overall, participants highlighted that understanding cause-and-effect and spatial-temporal interactions were key to physical reasoning, and noted that the benchmark items captured those aspects well.\footnote{Additional findings from this user study are discussed in section \ref{section:limitations}.} More details are in Appendix \ref{app:uxr}.

\section{Baseline Evaluations}

\label{section:eval}
CausalVQA is designed as a five-choice multiple choice test, with questions paired to reduce noise. The benchmark is subdivided into five question categories and three human-aligned difficulty levels. In what follows, we report zero-shot scores for the benchmark and each question subset using the \textit{paired} accuracy metric (where a model receives credit for a question only if both versions of the question are answered correctly). Empirically, models score $10$-$12\%$ lower on the paired version of the benchmark, indicating modest guessing and sensitivity to answer ordering and language (see Appendix \ref{app:unpaired}). To ensure reproducibility and facilitate participation in our benchmark leaderboard\footnote{\url{https://huggingface.co/spaces/facebook/physical_reasoning_leaderboard}}, we use \texttt{lmms-eval} \citep{lmms_eval2024}, a common open-source AI evaluation framework, along with the prompts and scoring function definitions included in our repository.

In addition to evaluating models on the full video and question-answer pairs, we report ablation results for two conditions in Appendix \ref{app:blind}: \textbf{blind}, where the model answers the question without seeing any part of the video, and \textbf{single-frame}, where the model answers the question given only the last frame in the video. 

We evaluate a range of contemporary state-of-the-art closed and open-weight models. The models vary in size and architecture, and we assess their performance across different benchmark dimensions.

\subsection{Human Baseline}

We measured human baseline performance by asking $273$ annotators (who had no previous exposure to the project) to answer benchmark questions. As mentioned previously, each item was completed by $15$ different annotators. Due to the cognitive load of answering all $1,786$ questions, each annotator only saw a subset of the questions, and no annotator saw both versions of the same question (human memory and consistency bias were a driving concern). To facilitate comparison with models, we provide bootstrap estimates (stratified by question category and difficulty) that use annotator responses as a proxy for human performance.  The results are presented in Table \ref{tab:performance}.

Overall, humans perform well, scoring $84.78\%$ on the test set and $\approx60\%$ on the hardest subset. Participants achieved the highest performance on \textit{descriptive} questions ($90.63\%$), while struggling more with \textit{counterfactual} ($76.57\%$) and \textit{hypothetical} ($75.92\%$) questions. On the \textit{Reasoning} portion of the benchmark (i.e. all question types excluding \textit{descriptive} questions), humans achieve $81.43\%$ accuracy.

\subsection{Models}
We evaluate a range of models on the CausalVQA benchmark, including:

\textbf{Locally-run open-weight models}: LLaVA-OneVision \citep{li2024llavaonevisioneasyvisualtask}, Qwen2.5VL \citep{bai2025qwen25vltechnicalreport}, PerceptionLM \citep{cho2025PerceptionLM}, and InternVL2.5 \citep{chen2025expandingperformanceboundariesopensource}. For each model family, we test a small, 7-8b parameter class version. Each was run on a single Nvidia A100 (80GB) GPU on our research cluster. Models are run with 16 uniformly sampled frames, which is a relatively high frame rate given the short duration of many videos. Temperatures are set to zero for reproducibility.

\textbf{Commercial closed models}: GPT-4o \citep{gpt4o} and Gemini 2.5 Flash \citep{gemini}. Both models are evaluated using the OpenAI API (Gemini's API did not work reliably for storing and sampling our clips) and 10 frames, uniformly sampled. Again, temperatures are set to zero. Gemini's `thinking' features are turned off, to give true zero-shot representation.

\subsection{Results \& Discussion}

\textbf{Overall Results}

At a high level, we see a meaningful gap between overall model performance and human accuracy across the benchmark tasks. Gemini had the highest performance with a $61.66\%$ paired score, which is $>22\%$ absolute lower than humans. On \textit{Reasoning} questions (which we define as all question types except \textit{descriptive}), Gemini still performed the best, but scored a more modest $53.46\%$, for a $\approx27\%$ gap to human performance.

\begin{table}[!htb]
    \centering
    \setlength{\tabcolsep}{2pt}
    \renewcommand{\arraystretch}{1}
    
    \small
    \begin{tabular}{llccccccccc}
        \toprule
        \multicolumn{3}{c}{} & \multicolumn{2}{c}{\textbf{Large/Closed}} & \multicolumn{4}{c}{\textbf{7-8b/Open}} & \multicolumn{1}{c}{} \\
        \cmidrule(lr){4-5} \cmidrule(lr){6-9}
        Category & 
        Difficulty & 
        N &
        \rotatebox{45}{GPT-4o} & 
        \rotatebox{45}{Gemini 2.5 Flash} & 
        \rotatebox{45}{InternVL2.5} & 
        \rotatebox{45}{LLaVa-OneVision} & 
        \rotatebox{45}{Perception-LM} & 
        \rotatebox{45}{Qwen2.5VL} & 
        \rotatebox{45}{Human} \\
        \midrule
        Anticipation & Easy & 110 & 49.09 & \textbf{55.45} & 53.64 & 38.18 & 47.27 & 46.36 & \textit{94.26} \\
        & Med & 64 & 28.12 & 32.81 & 35.94 & 28.12 & \textbf{42.19} & 26.56 & \textit{77.52} \\
        & Hard & 34 & 14.71 & \textbf{35.29} & 20.59 & 29.41 & 23.53 & 17.65 & \textit{60.17} \\
        \midrule
        Counterfactual & Easy & 28 & 64.29 & \textbf{85.71} & 50.00 & 57.14 & 60.71 & 53.57 & \textit{92.76} \\
        & Med & 34 & 44.12 & 35.29 & 38.24 & 35.29 & \textbf{47.06} & 41.18 & \textit{77.17} \\
        & Hard & 27 & 33.33 & 29.63 & 44.44 & \textbf{48.15} & 37.04 & 44.44 & \textit{58.96} \\
        \midrule
        Descriptive & Easy & 226 & 69.91 & \textbf{80.97} & 50.88 & 59.73 & 53.98 & 65.49 & \textit{95.96} \\
        & Med & 44 & 52.27 & \textbf{63.64} & 43.18 & 36.36 & 45.45 & 52.27 & \textit{76.19} \\
        & Hard & 18 & 27.78 & 44.44 & 27.78 & \textbf{44.44} & 38.89 & 38.89 & \textit{59.02} \\
        \midrule
        Hypothetical & Easy & 15 & \textbf{73.33} & 66.67 & 66.67 & 60.00 & 66.67 & 40.00 & \textit{92.63} \\
        & Med & 31 & 29.03 & \textbf{51.61} & 29.03 & 19.35 & 38.71 & 48.39 & \textit{77.32} \\
        & Hard & 21 & 14.29 & \textbf{42.86} & 38.10 & 4.76 & \textbf{42.86} & 28.57 & \textit{61.94} \\
        \midrule
        Planning & Easy & 75 & 65.33 & \textbf{70.67} & \textbf{70.67} & 52.00 & 69.33 & 57.33 & \textit{94.76} \\
        & Med & 42 & 42.86 & \textbf{73.81} & 57.14 & 59.52 & 61.90 & 47.62 & \textit{78.46} \\
        & Hard & 24 & 37.50 & \textbf{54.17} & 25.00 & 37.50 & 37.50 & 25.00 & \textit{60.27} \\
        \midrule
        \multicolumn{10}{c}{Aggregates} \\
        \midrule
        Anticipation & -- & 208 & 37.02 & \textbf{45.19} & 42.79 & 33.65 & 41.83 & 35.58 & \textit{83.54} \\
        Counterfactual & -- & 89 & 47.20 & \textbf{49.45} & 43.83 & 46.08 & 48.33 & 46.08 & \textit{76.57} \\
        Descriptive & -- & 288 & 64.58 & \textbf{76.04} & 48.26 & 55.20 & 51.73 & 61.80 & \textit{90.63} \\
        Hypothetical & -- & 67 & 34.32 & \textbf{52.23} & 40.29 & 23.88 & 46.26 & 40.29 & \textit{75.92} \\
        Planning & -- & 141 & 53.87 & \textbf{68.76} & 58.83 & 51.75 & 61.67 & 48.91 & \textit{83.99} \\
        \midrule
        Reasoning & -- & 505 & 43.16 & \textbf{53.46} & 47.12 & 39.60 & 49.10 & 41.78 & \textit{81.43} \\
        All & -- & 793 & 50.95 & \textbf{61.66} & 47.54 & 45.27 & 50.06 & 49.05 & \textit{84.78} \\
        \bottomrule
    \end{tabular}
    \caption{Performance metrics by model across benchmark question categories and difficulties. Reasoning questions refer to all question types except for \textit{descriptive}.}
    \label{tab:performance}
\end{table}

\textbf{Difficulty and Question Types}
Drilling down into specific subsets of the benchmark, we observe a few notable patterns. 
First, model performance generally aligns well with human-derived empirical difficulty levels, with models performing substantially worse on the \textit{hard} subset than the \textit{easy} one.

Second, performance across question types is uneven. There is generally a notable gap between models' performance on \textit{Reasoning} versus \textit{descriptive}  questions. For example, the top-performing model, Gemini 2.5 Flash, performed $53.46\%$ on \textit{Reasoning} questions compared to $76.04\%$ on \textit{descriptive} questions, whereas humans had a much smaller gap ($81.43\%$ vs $90.63\%$).

Anticipation questions, in particular, are difficult for this pool of models, while easy for humans: the average model here scores $39.34\%$ paired vs humans at $83.54\%$. Similarly, hypothetical questions were also difficult for the models, which averaged $39.55\%$ paired vs human at $75.92\%$. These results suggest that while SoTA models can approach human accuracy on understanding events that actually take place in the video (i.e. descriptive questions), there is still a substantial gap in models' ability to combine specific visual input with more abstract knowledge about the physical world to reason about what ``could have'' taken place or ``will likely'' take place given a particular world state and set of actions.

\textbf{Model-Specific Discussions}

Gemini 2.5 Flash was the best model we evaluated, with strong results across question categories. GPT-4o, in comparison, had its performance buoyed by its relative strength on easier descriptive questions. Among the open source models, PerceptionLM is extremely parameter efficient and has solid capabilities across anticipation, counterfactual, hypothetical, and planning tasks. In many situations, PerceptionLM was relatively better at the more difficult questions within those categories. Interestingly, InternVL2.5 and PerceptionLM both performed better on our reasoning aggregate than the much larger GPT-4o. 

\section{Related Work}

\begin{table*}[t]
    \centering
    \begin{NiceTabular}{lcccc}
        \toprule
        Benchmark & Realism & Diversity & Physical Causal Reasoning & Shortcut Mitigation \\
        \midrule
        CLEVRER (2020) & $\bullet$ $\circ$ $\circ$ & $\bullet$ $\circ$ $\circ$ & $\bullet$ $\bullet$ $\bullet$ & $\bullet$ $\bullet$ $\bullet$ \\
        ContPhy (2024) & $\bullet$ $\bullet$ $\circ$ & $\bullet$ $\bullet$ $\circ$ & $\bullet$ $\bullet$ $\bullet$ & $\bullet$ $\bullet$ $\bullet$ \\
        NExT-QA (2021) & $\bullet$ $\bullet$ $\bullet$ & $\bullet$ $\bullet$ $\bullet$ & $\bullet$ $\circ$ $\circ$ & $\bullet$ $\circ$ $\circ$ \\
        SUTD-TrafficQA (2021) & $\bullet$ $\bullet$ $\bullet$ & $\bullet$ $\bullet$ $\circ$ & $\bullet$ $\bullet$ $\circ$ & $\bullet$ $\bullet$ $\circ$ \\
        PerceptionTest (2023) & $\bullet$ $\bullet$ $\circ$ & $\bullet$ $\bullet$ $\bullet$ & $\bullet$ $\bullet$ $\circ$ & $\bullet$ $\circ$ $\circ$ \\
        ACQUIRED (2023) & $\bullet$ $\bullet$ $\bullet$ & $\bullet$ $\bullet$ $\circ$ & $\bullet$ $\bullet$ $\circ$ & $\bullet$ $\circ$ $\circ$ \\
        HourVideo (2024) & $\bullet$ $\bullet$ $\bullet$ & $\bullet$ $\bullet$ $\bullet$ & $\bullet$ $\bullet$ $\circ$ & $\bullet$ $\bullet$ $\circ$ \\
        \midrule
        CausalVQA (ours) & $\bullet$ $\bullet$ $\bullet$ & $\bullet$ $\bullet$ $\bullet$ & $\bullet$ $\bullet$ $\bullet$ & $\bullet$ $\bullet$ $\bullet$ \\
        \bottomrule
    \end{NiceTabular}
    \caption{Comparison of video reasoning benchmarks. The number of filled dots indicate the level of realism, diversity, physical causal reasoning, and shortcut mitigation present in the benchmark.}
    \label{table:demo}
\end{table*}

\label{section:related_work}
This section describes existing related benchmarks and contextualizes CausalVQA's contributions.
\subsection{Video Question Answering Benchmarks}
The universe of Video Question Answering (VQA) benchmarks is large \citep{zhong2022videoquestionansweringdatasets, li2024surveybenchmarksmultimodallarge}. Most, such as TGIF-QA \citep{Jang_2017_CVPR}, ActivityNet-QA  \citep{yu2019activitynetqadatasetunderstandingcomplex}, and TVQA \citep{lei2019tvqalocalizedcompositionalvideo}, tend to focus on descriptive questions, such as how many times an action happens, or the location of an object. Another cohort, such as NExT-QA \citep{xiao2021nextqanextphasequestionansweringexplaining}, PerceptionTest \citep{pătrăucean2023perceptiontestdiagnosticbenchmark}, and HourVideo \citep{chandrasegaran2024hourvideo1hourvideolanguageunderstanding}, include causal reasoning questions that tend to focus on probing goals and mental states (instead of physical principles), with few counterfactual items, or explore longform video understanding rather than assessing causal reasoning abilities at short horizons.

ACQUIRED \citep{wu2023acquireddatasetansweringcounterfactual} and Causal-VidQA \citep{li2022from} stand out as real-world video benchmarks with a specific focus on causal and counterfactual reasoning questions. However, relatively little was done to ensure visual groundedness in either benchmark. In contrast, CausalVQA was specifically designed with multiple safeguards to focus on questions that require a deeper causal understanding.

\subsection{Physical Reasoning Benchmarks}
Simulated benchmarks using controlled environments and compositions have been a popular solution to the problems presented by spurious cues in visual scenes and question language. PHYRE \citep{bakhtin2019phyrenewbenchmarkphysical} and IntPhys \citep{riochet2020intphysframeworkbenchmarkvisual} simulate physical scenes with visual primitives to test physical intuition. CLEVRER \citep{yi2020clevrercollisioneventsvideo}, ComPhy \citep{chen2022comphycompositionalphysicalreasoning}, and CRIPPVQA \citep{patel2022cripp} incorporated natural language to test causal reasoning in collisions between rigid objects. More recently, ContPhy \citep{zheng2024contphy} extended these tests to more diverse scenarios involving some non-rigid objects. While these benchmarks are effective diagnostic tests for causal reasoning, the simplification inherent in simulation limits realism and complexity. CausalVQA builds on the strengths of these systematic and controlled simulations while using actual video for improved real world generalization.

\subsection{Video Generation Benchmarks}
Another relevant category of benchmarks evaluates the physical commonsense of video generation models.
VideoPhy \citep{bansal2024videophyevaluatingphysicalcommonsense} and PhyGenBench \citep{meng2024worldsimulatorcraftingphysical} consist of text captions designed to probe text-to-video generation models' ability to produce videos that adhere to laws of physics, while Physics-IQ \citep{motamed2025generative} 
uses real video segments to prompt models to produce physically plausible continuations. While these benchmarks enable compelling demonstrations of the limitations of current video generation models, they are not designed to evaluate models' physical and causal understanding abilities in non-generative use cases.

\section{Limitations and Future Work}
\label{section:limitations}
CausalVQA introduced several features designed to ensure the benchmark accurately measures a model's understanding of video stimuli. However, the benchmark still exhibits some sensitivity to shortcuts, as a number of questions in our benchmark can be answered correctly when a strong model or human is given a single frame (see Appendix \ref{app:blind}). Much of this susceptibility is likely the result of correlations that govern real-world activities (e.g. the ending frame of a video sequence is usually predictive of the next set of actions, regardless of what happened prior to that frame), which are  difficult to remove from naturally occurring scenarios. Future work could explore more safeguards and filtering steps that further reduce the exploitability of such shortcuts.

Our benchmark focuses only on egocentric, goal-driven activities from the EgoExo4D dataset. Future work could explore more diverse sources of video data, and also incorporate more modalities of data, such as audio, to probe a richer multimodal experience.

In addition, our user research study highlighted that the ability to explain why an answer was chosen is a key expectation for `true' reasoning. Our benchmark does not yet incorporate explanations or thought traces to assess the process through which models arrive at an answer. Future work could incorporate more reasoning steps in the answer options to measure how well models are able to produce human-like thought processes.

 Finally, our extensive human baseline  presents an opportunity to better understand detailed correlations between human and model performance on individual items. As future work, we plan to further explore whether the errors models make are reflected in the disagreements seen in human responses (i.e., do both models and humans make similar mistakes). This will enable us to gain further insights on how humans and models differ in their reasoning strategies, which may inform future research directions to close the gap.

\section{Conclusion}
\label{section:conclusion}
We present CausalVQA, a carefully curated benchmark designed to evaluate multimodal models' causal reasoning skills in physically grounded scenarios in the real world. We built a hybrid human-and-model quality control loop to mitigate reliance on shortcuts. We also conducted in-depth human evaluation with over $273$ participants, whose answers we leveraged to segment the benchmark items into difficulty levels. The resulting dataset features $1,586$ items across \textit{counterfactual}, \textit{hypothetical}, \textit{anticipation}, \textit{planning}, and \textit{descriptive} question types, as well as \textit{easy}, \textit{medium}, and \textit{hard} levels. Zero-shot evaluation of six capable multimodal models shows that the best-performing model (Gemini 2.5 Flash) still performs meaningfully worse than humans on all question types and levels, but especially on items that require causal reasoning such as counterfactual questions and predicting what might happen next. Our aim for this work is to highlight this performance gap and encourage the development of advanced systems and architectures that truly understand the causal structure and physical principles underlying real-world activities and dynamics. We hope to establish CausalVQA as a benchmark challenge to help measure and stimulate the community's progress on this important research topic.

\section{Acknowledgments}
\label{section:ack}
We thank Michael Rabbat, Naila Murray, Claire Roberts, Michel Meyer, Lindsey Miller, Josh Terry, Carleigh Wood, Devansh Kukreja, Andrew Westbury, Roozbeh Mottaghi, Lorenzo Torresani, Xavi Puig, Kamila Benzina, and Rachel Kim for their contributions and support for the project.

\clearpage
\newpage
\bibliographystyle{plainnat}
\setcitestyle{numbers}
\bibliography{paper}

\clearpage
\newpage
\beginappendix

\section*{DataSheet}
CausalVQA benchmark datasheet, following the framework introduced by \citet{gebru2021datasheetsdatasets}.

\begin{longtable}{|p{6cm}|p{8cm}|}
    \toprule
    \multicolumn{2}{c}{\textsc{\textbf{Motivation}}} \\
    \midrule
    \textbf{For what purpose was the dataset created?} & CausalVQA is a video question answer (VQA) benchmark designed to test AI models on causal reasoning about videos. Typical video benchmarks have probed causality with simulations or have been broader video understanding tools; ours aims to combine real world situations with questions about causality.  \\ \midrule
    \textbf{Who created the dataset and on behalf of which entity?} & This dataset was created by the FAIR team at Meta. \\ \midrule
    \textbf{Who funded the creation of the dataset?} & Meta. \\ \midrule
    \multicolumn{2}{c}{\textsc{\textbf{Composition}}} \\ \midrule
    \textbf{What do the instances that comprise the dataset represent?} &  The instances are physical activity clips from the EgoExo4D video dataset with human annotated questions about causality.\\ \midrule
    \textbf{How many instances are there in total?} & 779 videos, 793 question pairs, 1586 individual question items. \\ \midrule
    \textbf{Does the dataset contain all possible instances or is it a sample (not necessarily random) of instances from a larger set?} & The clips are a human curated subset of the broader EgoExo4D dataset. \\ \midrule
    \textbf{Is there a label or target associated with each instance?} & The debug dataset includes targets. The test set has targets held back for use with our leaderboard. The dataset has the following metadata:\newline\newline
    \texttt{qid - a question identifier that gets used for pairing\newline 
            type - the question type (anticipation, counterfactual, descriptive, planning, hypothetical)\newline
            question - the text of the question\newline
            choices1 - the multiple choices\newline
            correct1 - the target for choices1 (removed from the test set)\newline
            choices2 - a perturbed and reordered set of multiple choices\newline
            correct2 - the target for choices2 (removed from the test set)\newline
            difficulty - the difficulty level from human baselines\newline
            renamed\_video - the videofile name}

    \\ \midrule
    \textbf{Is any information missing from individual instances?} & Only the targets (i.e. correct answers) in the held out set are not included.    \\ \midrule
    \textbf{Are relationships between individual instances made explicit?} & Yes, related questions (MCQ and MCQ`) have the same question id (qid). \\ \midrule
    \textbf{Are there recommended data splits?} & There are 2 different splits of the data. The debug set can be use to calibrate a model and set hyperparameters. The test set is the actual evaluation and targets are held out. Users can submit models' item-level responses to the leaderboard to determine performance. \\ \midrule
    \textbf{Are there any errors, sources of noise, or redundancies in the dataset?} & We have done multiple reviews of the dataset, both human and automated, but cannot guarantee that it is free of errors. All questions are offered as pairs of slightly linguistically perturbed, reordered answer sets. \\ \midrule
    \textbf{Is the dataset self-contained, or does it link to or otherwise rely on external resources?} & The dataset is self-contained, but is based on EgoExo4D and is available only through their license. \url{https://ego4ddataset.com/egoexo-license/}.    \\ \midrule
    \textbf{Does the dataset contain data that might be considered confidential?} & No, but it is bound by the EgoExo4D license. \\ \midrule
    \textbf{Does the dataset contain data that, if viewed directly, might be offensive, insulting, threatening, or might otherwise cause anxiety?} & No. \\ \midrule
    \multicolumn{2}{c}{\textsc{\textbf{Collection}}} \\ \midrule
    \textbf{How was the data associated with each instance acquired?} & The dataset is a subset of clips from EgoExo4D, a publicaly available dataset. Annotations were provided by contractors through a third party.\\ \midrule
    \textbf{What mechanisms or procedures were used to collect the data?} & Manual and automated (via Video Language Model, VLM) curation.    \\ \midrule
    \textbf{If the dataset is a sample from a larger set, what was the sampling strategy?} & It was less sampling and more curation for clips that would lend themselves to annotation about causality in the physical world. Individual items survived a multistep filtering and quality control process.   \\ \midrule
    \textbf{Who was involved in the data collection process and how were they compensated?} & The authors of this work and third party annotators hired through a vendor.   \\ \midrule
    \textbf{Over what timeframe was the data collected?} & The annotation data were collected between November 2024 and April 2025. Videos were previously collected by the Ego4D consortium. \\ \midrule
    \textbf{Were any ethical review processes conducted?} & Meta conducted an internal review with our legal and privacy teams prior to collecting the annotations (i.e. human-generated questions and answers) through the third-party vendor. \\ \midrule
    \textbf{Did you collect the data from the individuals in question directly, or obtain it via third parties or other sources (e.g., websites)?} & Videos - direct; Annotations - Third parties.   \\ \midrule
    \textbf{Were the individuals in question notified about the data collection? If so, please describe (or show with screenshots or other information) how notice was provided, and provide a link or other access point to, or otherwise reproduce, the exact language of the notification itself.} & All annotations were done under contract. Videos were previously collected by the Ego4D consortium and we sought permission for use as a member of the consortium. \\ \midrule
    \textbf{Did the individuals in question consent to the collection and use of their data? If so, please describe (or show with screenshots or other information) how consent was requested and provided, and provide a link or other access point to, or otherwise reproduce, the exact language to which the individuals consented.} & N/A. See above. \\ \midrule
    \textbf{If consent was obtained, were the consenting individuals provided with a mechanism to revoke their consent in the future or for certain uses? If so, please provide a description, as well as a link or other access point to the mechanism (if appropriate).} & N/A. See above. \\ \midrule
    \textbf{Has an analysis of the potential impact of the dataset and its use on data subjects (e.g., a data protection impact analysis) been conducted? If so, please provide a description of this analysis, including the outcomes, as well as a link or other access point to any supporting documentation.} & N/A. See above. \\ \midrule
    \multicolumn{2}{c}{\textsc{\textbf{Preprocessing}}} \\ \midrule
    \textbf{Was any preprocessing/cleaning/labeling of the data done?} & Videos were hand curated before being given to annotators. They were screened to focus on videos that contain substantial physical interactions.  \\ \midrule
    \textbf{Was the “raw” data saved in addition to the preprocessed/cleaned/labeled data?} & Raw data is available from EgoExo4D under their license. \\ \midrule
    \textbf{ Is the software that was used to preprocess/clean/label the data available?} & Prompts for the automated curation of questions and repair of annotation are provided in the appendices to this paper. \\ \midrule
    \multicolumn{2}{c}{\textsc{\textbf{Uses}}} \\ \midrule
    \textbf{Has the dataset been used for any tasks already?} & Yes, these data were used for the experiments that were presented in this paper.    \\ \midrule
    \textbf{Is there a repository that links to any or all papers or systems that use the dataset?} & No. \\ \midrule
    \textbf{What (other) tasks could the dataset be used for?} & No other tasks, only for model evaluation.\\ \midrule
    \textbf{Is there anything about the composition of the dataset or the way it was collected and preprocessed/cleaned/labeled that might impact future uses?} & No. \\ \midrule
    \textbf{Are there tasks for which the dataset should not be used?} & No. \\ \midrule
    \multicolumn{2}{c}{\textsc{\textbf{Distribution}}} \\ \midrule
    \textbf{Will the dataset be distributed to third parties outside of the entity on behalf of which the dataset was created?} & Yes, the dataset will be publicly distributed. \\ \midrule
    \textbf{How will the dataset will be distributed?} & Via the Ego4D AWS S3 instance after signing their license. \\ \midrule
    \textbf{Will the dataset be distributed under a copyright or other intellectual property (IP) license, and/or under applicable terms of use (ToU)?} & The license of the dataset is \textbf{cc-by-nc}. Videos are distributed under the EgoExo4D license. \\ \midrule
    \textbf{Have any third parties imposed IP-based or other restrictions on the data associated with the instances?} & See Ego4D license restrictions. \\ \midrule
    \textbf{Do any export controls or other regulatory restrictions apply to the dataset or to individual instances?} & N/A \\ \midrule
    \multicolumn{2}{c}{\textsc{\textbf{Maintenance}}} \\ \midrule
    \textbf{Who will be supporting/hosting/maintaining the dataset?} & Meta. \\ \midrule
    \textbf{How can the owner/curator/manager of the dataset be contacted?} & Please contact the corresponding author of this paper. \\ \midrule
    \textbf{Is there an erratum?} & No. \\ \midrule
    \textbf{Will the dataset be updated?} & Uncertain. The dataset may be updated by versioning. \\ \midrule
    \textbf{If the dataset relates to people, are there applicable limits on the retention of the data associated with the instances (e.g., were the individuals in question told that their data would be retained for a fixed period of time and then deleted)? If so, please describe these limits and explain how they will be enforced.} & N/A. \\ \midrule
    \textbf{Will older versions of the dataset continue to be supported/hosted/maintained? If so, please describe how. If not, please describe how its obsolescence will be communicated to dataset consumers.} & Uncertain. We are not planning to update it at this point. \\ \midrule
    \textbf{If others want to extend/augment/build on/contribute to the dataset, is there a mechanism for them to do so?} & No. \\ \bottomrule
    \caption{Datasheet for CausalVQA.}
    \label{tab:datasheet}
\end{longtable}

\clearpage
\newpage

\section{Human Tasks}

This section details all tasks with a human dependence for either benchmark construction of baselining/difficulty assignment. Headings relate to the specific parts of the paper body where they apply.

\subsection{Question Generation Guidelines}
\label{question_guidelines}
The following is the set of guidelines we developed with our third party annotators to generate questions from the set of EgoExo4D videos used for the benchmark. Some elements have been removed for brevity, but the essential sections are preserved. Sixteen annotators spent over $300$ hours on this task; each clip cost $\$52$ to review and generate questions.

\textbf{Overview}

This document describes the process human annotators should use for question creation. We are building a novel benchmark to measure AI progress in reasoning about the physical world. This particular portion of the benchmark will rely on testing in-depth understanding of real videos, formulated as a multiple choice question task.

The annotator’s job will be to watch video clips and devise questions that are obvious to a human and supply a ground truth response.

Each clip should be reviewed by three annotators and each annotator should seek to create around five questions per suitable clip (instructions on determining suitability below).

\textbf{Question Types}

We are primarily focused on what we call ‘reasoning’ questions: anticipation (prediction), planning, counterfactual, and hypothetical. In addition, we are tangentially interested in understanding the facts in a video (we call these descriptive questions). A definition of each question is given briefly below. Further task detail and good/bad examples will be provided in Examples (omitted for brevity). 

\textit{Descriptive}: descriptive questions should test for understanding the facts presented in the video (e.g. actions performed, temporal order, the goal the person is trying to achieve, etc). Descriptive questions are not the point of the benchmark, they mostly supply additional context for the other question types (e.g, a foundation that identifies the action that a human is performing in a clip narrows the possible paths the clip may take, making it more deterministic and allowing for more probing questions of other types). Descriptive questions include sequences if the clip contains all the information and the question asks only for a regurgitation of the steps – do not confuse this with planning (which requires understanding a goal and making inferences about a path to that goal).

\textit{Anticipation}: anticipation questions test for the ability to make intuitive predictions about what is likely to happen next in a video (e.g., simple physics, like the trajectory of a ball). Designing good anticipation questions is somewhat more difficult than descriptive questions, as the thing being predicted should feel deterministic to a human without being so simplistic that it could be guessed without understanding the clip.

\textit{Planning}: planning questions are about achieving some set objective. They require understanding sequences of linked events (e.g., building something, where subsequent steps depend on foundational steps, like stacking blocks). Questions here may test the understanding of the sequence itself, or a rationale for why a human in a clip may perform an action (i.e., they must perform the action to advance towards their goal).

\textit{Counterfactual}: counterfactual questions assess broad understanding of the situation and to think about possibilities paths that depart from the specific facts in a clip. These can usually be framed as ‘what if…’ scenarios that were not observed (e.g., what would have happened next if a person moved in a different direction from the one observed in the clip).

\textit{Hypothetical}: hypothetical questions are also designed to assess broader understanding of the situation and should use elements of the clip as a starting point and typically create a ‘what else…’ imagined scenario (e.g., what else would make sense from the point of view in the clip even if it departs from the specifics of the clip)

\textbf{Guidelines and Steps for Question Creation:}

\begin{itemize}

\item \textbf{Step 0:} Annotators were asked to familiarize themselves with the task and common pitfalls (omitted, again) in video review.

The first step is to scan the clip to see if it is suitable. We gave the following helpful heuristic to determine if a clip may work:

(if yes, proceed):
Is it easy for an average adult to understand what the video is about? 
(at least one yes): 

AND Does the activity in the clip have intuitive physical motions?
         
         OR Does the activity in the clip have some sort of goal or direction?
         
         OR Does the activity in the clip have multiple steps?

\item \textbf{Step 1:} The annotator scans the full clip. Scanning the clip helps the question creator assess whether the current clip will work for questions. Generally, the creator should watch with the intent of creating several questions of different types. However, not all clips are suitable (some activities may not be sufficiently deterministic for prediction questions or may be filled with arbitrary activity). A quick scan of the clip will help the overall speed and integrity of the process. If the clip is not suitable for the question generation task, we ask the annotator to record the reason why. We remind the annotator that they now know what will happen in the clip and to be certain that the questions they create do not assume future knowledge.

\item \textbf{Step 2:} The annotator generates a single descriptive question as the basis for understanding the clip (this is a short segment that contains the minimum one would need to recognize the activity in the clip and will serve as a basis for additional inferences made in more complex questions). Then creates a suggested answer. He/she records the timestamp needed to frame the question (a beginning and end).

\item \textbf{Step 3:} Next the annotator identifies a segment for a followup question of a more complex type (anticipation, planning, counterfactual, hypothetical, etc) and generate a new question. Then adds a suggested answer.

\item \textbf{Step 4:} Repeats step 3 to generate additional questions (anticipation, planning, counterfactual, hypothetical, etc) until out of obvious ideas

\item \textbf{Step 5:} When completed with the initial pass at making questions and answers, the annotator should review to validate general correctness. He/she should assess that the questions and answers are not subject to any common pitfalls. (Note: checklists for validation and pitfalls removed for brevity)

\end{itemize}

\subsection{Human Quality Assurance Guidelines}

The following is the set of guidelines we developed for human quality assurance of \texttt{MCQ} and \texttt{MCQ'}. 

\textbf{Context}
We are building a new benchmark dataset that measures a model’s ability to understand and reason about the physical world through multiple choice questions (MCQs) about videos. 

\textit{In Phase 1: Question Creation}, we asked annotators to generate question and answer pairs about a set of video clips. Each video clip contains an egocentric view of a person performing a specific activity (e.g. cooking, playing sports, fixing something, etc). The annotators in Phase 1 then created various types of questions about each clip and provided the corresponding “correct” answers. Annotators also identified the specific start and end time stamps that are relevant for each question.

\textit{In Phase 2: Quality Assurance}, we would like annotators to review various components of the items created in Phase 1. These components include the question, the correct answer, alternative answers (i.e. “distractors”), begin/end time stamps, and question type labels. The purpose of the Quality Assurance step is to remove or fix any errors in order to ensure that the final set of questions is as clear and unambiguous as possible. 

Below are guidelines for completing the Quality Assurance task.

\newpage
\textbf{Workflow}
\begin{enumerate}
    \item Annotator watches the video clip (segmented at the start/end time stamps provided by Phase 1 annotators).
    \item Reads the question and set of choices (without seeing which one is the correct answer)
    \item Attempts to select the correct answer
    \item Tool reveals the correct answer
    \item Answers the following questions (more than one item may be wrong - several of them may need to be flagged):
    \begin{enumerate}
        \item According to the definitions, what type of question is this? 
    \begin{enumerate}
        \item Descriptive
        \item Anticipation
        \item Planning
        \item Counterfactual
        \item Hypothetical
    \end{enumerate}
    \item Is there sufficient information in the video to answer the question correctly?
    \begin{enumerate}
        \item Yes - the video contains the appropriate amount of information
        \item No - the video ends too early; not enough information is shown to answer the question
        \begin{enumerate}
            \item Please supply a better end time stamp
        \end{enumerate}
        \item No - the video starts too late; not enough information is shown to answer the question
        \begin{enumerate}
            \item Please supply a better start time stamp
        \end{enumerate}
    \end{enumerate}
    \item Do you need to watch the video in order to choose the correct answer?
    \begin{enumerate}
        \item Yes - you need to watch the video in order to answer correctly
        \item No - it’s easy to guess the correct answer without watching the video
        \begin{enumerate}
            \item What about the question or answers makes it easy? Please be specific
        \end{enumerate}
    \end{enumerate}
    \item Is the question clear and well-written?
    \begin{enumerate}
        \item Yes, the question is clear and well-written
        \item No, the question is confusing or poorly written
        \begin{enumerate}
            \item Make edits to improve the question
        \end{enumerate}
    \end{enumerate}
    \item Do the answer options make sense, and is there only one answer that is most correct?
    \begin{enumerate}
        \item Yes, all of the answer options make sense, and there is only one answer that is most correct
        \item No, some of the answer options are equally “correct”
        \begin{enumerate}
            \item Select the answer options that are equally correct
            \item Make edits to improve the options such that there is only one correct answer
        \end{enumerate}
        \item No, some of the answer options are confusing or poorly written
        \begin{enumerate}
            \item Select the answer options that are confusing or poorly written
            \item Make edits to improve the options
        \end{enumerate}
    \end{enumerate}
    \end{enumerate}
\end{enumerate}

\subsection{Human-Derived Empirical Difficulty Levels and Human Baselining Guidelines}

We conducted a robust human baseline for CausalVQA with $273$ annotators in order to provide a more clear and interpretable reference against which benchmarked models can be measured. As mentioned in the paper body, human baselining was performed only on filtered questions to reduce cases where models could respond correctly without visual understanding.

Our baseline had the following criteria and process:
\begin{enumerate}
    \item A census-matched pool of age and gender for hired evaluators.
    \item Multiple questions included solely to filter for both attention and task understanding.
    \item To account for variance in the length of videos associated with each question, we standardize the amount of time an evaluator will spend across questions by assigning a set of videos with an approximately equal absolute duration per evaluator.
    \item Evaluators are never shown both possible questions for a video, to eliminate the influence of within-question bias and memory on selected answers.
    \item Task descriptions given to evaluators emphasize in multiple ways the goal of e.g. predicting a sequence or end state. Our goal was to ensure strong understanding by the evaluators (even though none had prior exposure to the benchmark), such that the baseline would more closely represent performance of these tasks in the wild when a person’s intent is to arrive at some conclusion informed by their understanding of the world.
\end{enumerate}

\subsection{Qualitative Assessment Guidelines}
\label{app:uxr}
We ran a user research study that consisted of twelve 90 minute remote qualitative interviews in the US, Canada, \& UK in early 2025 that went in depth with participants on their definitions of ‘reasoning’ and ‘problem solving’. This qualitative research involved members of the general population balanced by age and gender who self-reported to have some knowledge and experience with AI products. Participants were asked to review the work-in-progress items from the benchmark to assess how well these map to their own definitions of reasoning.

\textbf{Goals of User Research}

\begin{itemize}
    \item How do general population audiences define and conceptualise the concepts ‘reasoning’ \& different reasoning modalities (reasoning about images, videos, physical 3D spaces)?
    \item What are participants' thoughts on what types of general reasoning tasks/ capabilities are most important for AI models to ‘get right’?
    \item How do different audiences react to the proposed CausalVQA benchmark tasks? Do they feel that they effectively test reasoning? If a model passed these tests would people agree that AI could effectively ‘reason’? (e.g. if a model got all the questions correct, does that make people feel satisfied that the model can reason?)
    \item Do participants care more about the capability definition of ‘reasoning’ vs the usefulness of the model?
\end{itemize}

\textbf{Interview Script Summary}

\textbf{Section 1: }Definitions\textbf{: }We briefly ask participants to introduce themselves \& how they define and understand concepts of reasoning. We ask participants about the concepts ‘reasoning’, ‘core learning’, ‘logic’ and how they think this can be applied in AI.

\textbf{Section 2: }Benchmark Evaluations

We show participants the CausalVQA benchmark to understand if they feel that they effectively test physical world reasoning and why.

\begin{itemize}
    \item Participant to run through task
    \item Speak aloud how they do the task
    \item Review tasks and see if this aligns with their definition of reasoning.
    \item Ask them what, if possible, would be a better test physical world ‘reasoning’
\end{itemize}

\section{Prompt Details for Question Creation Pipeline}

This appendix covers details of our automated question creation and quality assurance pipeline. Prompts are provided in the order executed and headings reference the section of the paper where the prompts apply. Occasional output examples are spaced throughout to show what, specifically, the pipeline does.

\subsection{Distractor Generation}

\promptbox{Prompt for VLM to to generate initial answer given video and question}{You will be given a question to answer. Please answer it in a single sentence that uses common sense, grammatically and idiomatically correct writing. Then, write a rationale for your answer, again in a single sentence, but make certain to keep it visually grounded in the clip.

Answers will follow this format:

-----

Answer: "your answer" END

Rationale: "the rationale for the answer" END

-----
Here is an example:

Question: What will the person do with the knife?

Answer: "The person will cut the tomato." END

Rationale: "The person placed the tomato on the cutting board and their hand is moving towards the tomato while holding the knife." END

Now, here is the prompt you must respond to:

Question: {question}

Remember to format your answer correctly. That is a common mistake you have made in the past. Sometimes you also forget the line terminator 'END'. Do not forget to follow your Answer and Rationale with 'END'.

Do not repeat yourself or the prompt. Do not give extraneous information other than the two elements of the answer.}

\promptbox{Prompt for VLM to generate descriptons of clips}{You will be given a clip to view. Please describe it at a level sufficient for someone who is not able to see the clip to understand its contnet. Pay special attention to the activities of people and the mechanics/physics of the things with which the people interact, as well as the things themselves. You do not care about the atmosphere, only the people/objects/activities (you do care about the specific people, objects, and activities, so try to correctly name them rather than using a generic term). Keep it a single paragraph and do not repeat yourself.
\newline

Avoid the following mistakes: 1. multiple paragraphs, 2. redundant, irrelevant information. 3. reporting on lighting, atmosphere, or mental states. 4. Making assumptions about the specific roles/jobs/titles of people when that information is not explicitly given.}

\newpage

\promptbox{Prompt for VLM to generate seed distractors}{You are an expert at testing other AI models by generating distracting multiple choice answers. You follow the following steps for this task:

1. Carefully read and understand the Revised Question and the Revised Answer.

2. Carefully observe the Clip given with the prompt.

3. Carefully read and understand the Scene Description and the Next Scene Description.

4. Devise a list of four Distractors that are extremely plausible and could be correct answers to the Revised Question. The answers should extremely plausible, but incorrect, based on what you see in the Clip . They could be correct given the information in the Scene Description and Next Scene Description, but are not. The Distractors must be **extremely similar** in length, specificity, and complexity to the Revised Answer. Distractors should not simply invert or revise adjectives or substitute synonyms if a better answer is possible while staying focused on what was present in the Clip. Distractors should also be **distinctly different** from each other and from the Revised Answer. Consider alternative perspectives, assumptions, or interpretations that could lead to different conclusions. Do not mention information that could not be known at the time (e.g., things that are unique to the Next Scene Description that don't appear in the Scene Description). **Never** give a list of Distractors that contains multiple elements that are the same.

5. Check to make the Distractors are similar to, but differentiated from the Revised Answer. Distractors **must** be similar in length and complexity to the Revised Answer. Distractors **must** have unique meanings when compared to each other and the Revised Answer, but rely on similar elements. Make certain that objects, people, and activities in the Distractors rely on what you see in the Clip and specific language from Scene Description and Next Scene Description and the Revised Answer. Modify the Distractors as needed unil you have your list of four that are different from each other and the Revised Answer.

6. Ensure that the Distractors make sense and are extremely plausible as answers to the Revised Question given what you saw in the Clip, even though they are incorrect. Revise the Distractors as needed unil you have your list of four.

7. Check to make certain you have not included the Revised Answer (or its equivalent) among the Distractors. Modify the Distractors as needed to ensure they are all different from the Revised Answer.

8. Supply the rationale for your Distractors.

Responses must follow this format specifically:

-----

Distractors: "[Distractor 1| Distractor 2| etc...]" END

Rationale: "the rationale for the Distractors." END
-----

You often send the Revised Answer in the Distractors. Follow these steps to repair this error:

1. Before you output the Distractors, check to see if the Revised Answer is present in the Distractors (remember, they are separated by |).

2. If the Revised Answer is present, remove it from the Distractors and generate another Distractor that is different from the Revised answer and the remaining Distractors, but similar in length, language, and complexity.

3. Return the new Distractors and Rationale in the correct format.

-----

Here is an example of correct format:

Distractors: "['distractor 1.'| 'distractor 2.'| 'distractor 3.'| 'distractor 4.']"\newline

Rationale: "The original answer is correct because...The other answers are incorrect because..."

-----

Important:

Please double check to ensure your response fits the requested format. Missing the END line terminator is a common mistake you have made in the past. Another common mistake is not returning the Distractors as a list separated by a pipe |. A final common mistake is not returning the Rationale as a single string. Additionally, make certain you return your work and not one of the examples.

Using the above context and instructions, here is the question you need to answer:

Revised Question: question

Revised Answer: answer

Scene Description: scenedesc

Next Scene Description: nextscenedesc

Do not repeat yourself or the prompt. Do not give extraneous information other than the two elements of the answer.}

\begin{table}[!h]
  \centering
  \caption{Example question-answer and associated seed distractors from the initial steps in the pipeline.}
  \begin{tabular}{ll}
  \toprule
    Question & Which way will the person turn next? \\
    \midrule
    Answer & The person will turn to the right next \\
    \midrule
    Seed Distractors & 
      \begin{tabular}{@{}l@{}}
        1. The person will turn to the left next \\
        2. The person will turn to the left and then to the right \\
        3. The person will turn to the right and then to the left \\
        4. The person will turn to the left and then to the left again
      \end{tabular} \\
    \bottomrule
  \end{tabular}
  
  \label{fig:annotator-question-answer}
\end{table}
\vspace{25pt}

\subsection{Language Refinement}
\label{app:language_refinement}

\promptbox{Prompt for LLM to refine the questions and answer optons}{You are an expert at understanding questions and editing them to use common sense language. You follow the following steps in this task: 

1. Carefully read the Original Question and the Original Answer. 

2. Give a one sentence rewritten version of the answer that improves it by using simple, concise (substitute one word for two, if possible), common sense, idiomatically correct American English in active voice. Please sound as polished as possible and make certain to capture the gist of the Original Answer. Please change answers to binary questions a non-binary equivalent.

3. Check that the rewritten answer makes sense for the question. 

4. Ensure the rewritten answer captures the gist of the Original Answer. 

5. Next, try to rewrite the question using normal, common sense, idiomatically correct American English in active voice. Ensure that language in the question reflects the language in the rewritten answer.

6. Revise the rewritten question and substitute any words that may provide a hint at the answer (e.g., specific nouns or verbs) with generic/neutral ones. Now make certain that the rewritten question has the same meaning as the Original Question.

7. Supply your rationale for the changes you make in Answer Reasons and Question Reasons, respectively.

Responses will follow this format:

-----

A. \texttt{Rewritten Answer}: "the rewritten answer you came up with in step 2, adjusted for any changes from steps 3 and 4." END

B. \texttt{Rewritten Question}: "the rewritten question you came up with in step 5, adjusted for any changes from step 6." END

C. \texttt{Answer Reasons}: "the rationale for the changes you made between the Original Answer and the Rewritten Answer." END

D. \texttt{Question Reasons}: "the rationale for the changes you made between the Original Question and the Rewritten Question." END

$-----$

Remember to include only parts A-D in your answer in the correct format. 

Here is an example:

Original Question: What would happen if the person doesn't stretch out their right foot?

Original Answer: The person stretches out their right foot to return the ball through the yellow cones. If the person doesn\'t stretch to out their foot, the ball would not be returned through the yellow cones.

Rewritten Answer: They could not return the ball through the yellow cones if they failed to extend their right foot. 
Rewritten Question: What would happen if the person failed to extend their right foot?
\newline

Now, using the above context and instructions, here is the real material you need to edit:

Original Question: {question}

Original Answer: {answer}
\newline

Remember to include only parts A-D in your answer in the correct format. Do not repeat yourself or the prompt. Do not give extraneous information other than items A-D.}

\promptbox{Prompt for LLM to check meaning is preserved}{You are an expert at performing comparisons between text and editing them so that they are similar. You follow these steps in this task: 

1. Carefully read the Original Answer and the New Answer.

2. Assess whether or not the New Answer is very similar to the Original Answer. Do they have the same essential meaning? If the New Answer adds information not present in the Original Answer that changes the meaning, correct the New Answer so that captures the gist of the Original Answer. If the New Answer is missing critical physical/mechanical detail that a blind person would need to understand the main action of the scene, please capture those missing elements from the Original Answer. Always attempt to improve on the conciseness, grammer, style and phrasing. The result will be correct and have removed extraneous bits that don't contribute to meaning. Use the Original Answer if it is better than any other you can devise.

3. Next, carefully read the Original Question and the New Question.

4. Assess whether or not the New Question is very similar to the Original Question. Do they have the same essential meaning? If the New Question adds information not present in the Original Question or changes the meaning, correct the New Question so that it captures the gist of the Original Question, while improving on the grammer and phrasing. 
\newline

Output must follow this format:

-----

A. Rewritten Answer: "the rewritten New Answer you came up with in step 2" END

B. Rewritten Question: "the rewritten New Question you came up with in step 4" END

C. Answer Reasons: "the rationale for the changes you made between the New Answer and the Rewritten Answer." END

D. Question Reasons: "the rationale for the changes you made between the New Question and the Rewritten Question." END

-----
\newline

Here is an example:

New Answer: The person uses their left foot to stop the ball.
Original Answer: The person reaches out with their left foot to stop the ball from moving. They stop the ball with their left foot when they are near the right cone.

Discussion: The New Answer captured the important information in the Original Answer in a concise manner without adding anything to it. It removed extraneous detail, while keeping the language concise and gramatically/idiomatically correct.

Rewritten Answer: The person extends their left foot to stop the ball.

Reason: The Rewritten Answer captures the same information as the New Answer and Original answer, but uses a more idiomatically normal sentence.

New Question: Why does the person use their left foot to stop the ball?

Original Question: Why does the person reach out with their left foot?

Discussion: The New Question does a good job of using common English phrasing, but changes the nature of what is being asked. It needs to be edited to have the same meaning as the original, while keeping the language concise and gramatically/idiomatically correct.

Rewritten Question: Why does the person extend their left foot?

Reason: The Rewritten Question corrects the New Answer by removing the introduced information about the ball stopping and introduces more idiomatically correct language.
\newline

Here is a second example:

New Answer: The person uses the same foot to stop and then kick the ball back to the goalkeeper.

Original Answer: The person is using the same foot they stop the ball with the then kick the ball back to the goalkeeper.

Discussion: The New Answer captured the important information in the Original Answer in a concise manner, but doesn't answer the Original Question. Logical detail should be added, while keeping the language concise and gramatically/idiomatically correct.

Rewritten Answer: The person will kick with their left since they are stopping and kicking the ball back to the goalkeeper with the same foot.

Reason: The Rewritten Answer fixes the New Answer by giving enough detail for a logically known answer to the Original Question and keeps the language concise and gramatically/idiomatically correct.

New Question: Why does the person use their right foot to kick the ball back to the goalkeeper?

Original Question: Why wouldn't the person kick the ball back with their right foot?

Discussion: The New Question changes the nature of what is being asked substantially, confusing whether or not a person would or wouldn't use their right foot. It needs to be edited to have the same meaning as the original, while keeping the language concise and gramatically/idiomatically correct.

Rewritten Question: Why won't the person kick the ball with their right foot?

Reason: The Rewritten Question fixes the New Answer by keeping the meaning consistent with the Original Question, while improving the fluency of the language.

-----

Now, using the above context and instructions, here is the real material you need to edit:

New Answer: {newanswer1}

Original Answer: {answer}

New Question: {newquestion1}

Original Question: {question}
\newline

Remember, you must only provide items A-D in the correct format. Do not repeat yourself or the prompt. Do not give extraneous information other than items A-D.}

\promptbox{Prompt for removing hint words}{You are an expert at revising question language to remove hints at the answer. You follow the following steps for this task:

1. Carefully read the Original Question and the Original Answer and assess if any Important Words in the Original Question give a hint as to what the Original Answer may be. Important Words will only be found in the Original Question. There can be any number of Important Words in your Key Words list.

2. Carefully read the AI First Pass Answer and the Original Answer and give the similarity a Score from 0 (not similar) to 2 (essentially the same).

3. If your Score is equal to or greater than 1, assess if any Important Words in the Original Question give a hint as to what the AI First Pass Answer may be. Important Words will only be found in the Original Question. Cross check it against your Key Words list and add any new Important Words to the Key Words list.

4. Revise the Original Question into a Rewritten Question that substitutes the Key Words for neutral/generic words you can think of so that it does not provide a hint as to the Original Answer or AI First Pass Answer, but would keep the question's meaning extremely similar. Be careful not to change the meaning of the Original Question (e.g., substituting "defensive position" for "behind the net") 

5. Ensure that the Original Answer would still make sense as a correct response to the Rewritten Question. 

6. Fix the Rewritten Question using highly generic and neutral language that maintains the Original Question's meaning so that the Original Answer will make sense.

7. Supply the rationale for your changes.

Responses must follow this format specifically:

-----

A. Rewritten Question: "the answer you came up with in step 1, 2, 3, 4, and 6." END

B. Similarity Score: "the Score from step 2." END

C. Key Words: "[Important Word 1, Important Word 2, etc...]" END

D. Question Reason: "the rationale for the changes in the Rewritten Question." END

-----

Please double check to ensure your response fits the requested format. Missing the END line terminator is a common mistake you have made in the past. Another common mistake is not returning the Key Words as a list, even if it is a single element or there are no Important Words.
\newline

Using the above context and instructions, here is the question you need to answer:

Original Question: {newquestion2}

Original Answer: {newanswer2}

AI First Pass Answer: {rawanswer}

Remember, you must provide items A-D in the correct format. This is extremely important for the goodness of the world. Do not repeat yourself or the prompt. Do not give extraneous information other than items A-D.}

\newpage
\subsection{Language Purtubation and Reordering}

\promptbox{Prompt to generate perturbed distractors}{You are an expert at testing other AI models by generating distracting multiple choice answers. You follow the following steps for this task:
\newline

1. Carefully read and understand the Revised Question and the Revised Answer.

2. Carefully read and understand the Possibilities. These are potential answers to the Revised Question that are often incorrect. They are a list of four sentences (usually separated by "|", ",", "." or " ") and are sometimes prefaced by "Distractors: ". These have been generated by a model that can understand video clips and are designed to be incorrect, but that model sometimes includes an equivalent to the Revised Answer. Remove these Revised Answer equivalents from consideration of the Possibilities. Heavily rely on the remaining Possibilities.

3. Carefully read and understand the Scene Description and the Next Scene Description.

4. Devise a list of four New Distractors that are polished versions of the remaining Possibilities and that are plausible and could be correct answers to the Revised Question. They **must** be similar in length and complexity to the Revised Answer. The answers should be extremely plausible and could be correct given the information in the Scene Description and Next Scene Description. However, they should also be **distinctly different** from each other and from the Revised Answer. Consider alternative perspectives, assumptions, or interpretations that could lead to different conclusions. Do not mention information that could not be known at the time (e.g., things that are unique to the Next Scene Description that don't appear in the Scene Description).

5. Check to make sure the New Distractors are not only **completely** plausible but also **divergent** from each other and from the Revised Answer. Encourage yourself to think outside the box and consider unconventional explanations or scenarios. Double check that you are thinking of different New distractors - **do not repeate** the same one multiple times. **Never do anything like this** these New Distractors are not different: The person turns to the right to see where the ball is going after bouncing off the rim| The person turns to the right to see where the ball is going after bouncing off the rim| The person turns to the right to see where the ball is going after bouncing off the rim| The person turns to the right to see where the ball is going after bouncing off the rim

6. Ensure that the New Distractors make sense as answers to the Revised Question in the same context, even though they are incorrect. **Never** repeate the Revised Answer in the New Distractors. **Never** repeate any of the individual elements in the New Distractors list. Revise the New Distractors as needed unil you have your list of four.

7. Supply the rationale for your New Distractors.

8. Generate a second list of Synonym Distractors that substitute common sense (not overly stilted) synonyms for some words in the Distractors (you'll want to change more than one, generally), but keeps the meaning of the Distractors the same.

Responses must follow this format specifically:

-----

A. New Distractors: "[New Distractor 1| New Distractor 2| etc...]" END

B. Synonym Distractors: "[Synonym Distractor 1| Synonym Distractor 2| etc...]" END

C. Rationale: "the rationale for the Distractors." END

-----

Please double check to ensure your response fits the requested format. Missing the END line terminator is a common mistake you have made in the past. Another common mistake is not returning the Distractors as a list separated by a pipe |. A final common mistake is not returning the Rationale as a single string.
\newline

Using the above context and instructions, here is the question you need to answer:
Revised Question: {newquestion3}
Revised Answer: {newanswer2}
Possibilities: {distractors}
Scene Description: {scenedesc}
Next Scene Description: {nextscenedesc}

Remember, you must provide only items A-C in the correct format. Do not repeat yourself or elements of the prompt. Do not give extraneous information other than items A-C.}

\begin{table}[!h]
  \centering
  \caption{The result of the language refinement process showing increased diversity and a correction to item 3, which could have been an alternative correct answer (see prior example)}
  \small 
  \begin{tabular}{p{0.5\textwidth}p{0.5\textwidth}}
    \toprule
    \textbf{Seed Distractors} & \textbf{Refined Distractors} \\
    \midrule
    \begin{minipage}[t]{\linewidth}
      \begin{enumerate}
  \item The person will turn to the left next
  \item The person will turn to the left and then to the right
  \item The person will turn to the right and then to the left
  \item The person will turn to the left and then to the left again
\end{enumerate}
    \end{minipage} &
    \begin{minipage}[t]{\linewidth}
      \begin{itemize}
      \item The person will advance to strike the ball  
      \item The person will retreat to gain distance 
      \item The person will pivot to the left to evade a challenger
      \item The person will leap up to meet the ball
      \end{itemize}
    \end{minipage} \\
    \bottomrule
  \end{tabular}

\end{table}

\begin{table}[h]
\caption{Example showing the perturbed and reordered multiple choice answer sets}
  \centering
  \small
  \begin{tabular}{p{0.5\textwidth}p{0.5\textwidth}}
    \toprule
    \textbf{\texttt{MCQ}} & \textbf{\texttt{MCQ'}} \\
    \midrule
    \begin{minipage}[t]{\linewidth}
    
  A. The person will move backward to create space
  
  B. The person will turn to the right next

  C. The person will move forward to kick the ball
  
  D. The person will turn to the left to avoid an opponent
  
  E. The person will jump up to head the ball
    \end{minipage} &
    \begin{minipage}[t]{\linewidth}
    A. The person will turn to the right next
    
    B. The person will leap up to meet the ball
    
    C. The person will retreat to gain distance
    
    D. The person will advance to strike the ball

    E. The person will pivot to the left to evade a challenger
    \end{minipage} \\
    \bottomrule
  \end{tabular}

\end{table}
\subsection{Ensuring Visual Groundedness}

\promptbox{Prompt for LLM to remove questions answerable without visual information}{You are an expert at answering multiple choice questions. You follow the following steps for this task:
\newline

1. Read the Question.

2. Read the list of Possible Answers. It is supplied as a string of numbered answers separated by a pipe |.

3. Choose Your Answer from the Possible Answers -- pick the one that is most correct. You must choose an answer, even if you are not sure.

4. Return Your Answer as a single string. Record Your Answer Number. Supply the Rationale for your answer.
\newline

Responses must follow this format specifically:

-----

A. Your Answer: "Your Answer from the Possible Answers as a string" END

B. Your Answer Number: "The number at the beginning of Your Answer returned as an integer" END

C. Rationale: "the rationale for Your Answer" END

-----

Please double check to ensure your response fits the requested format. Missing the END line terminator is a common mistake you have made in the past. Another common mistake is not returning Your Answer or Your Answer Number in the correct format. A final common mistake is not choosing to answer a question.
\newline

Using the above context and instructions, here is the material:
Question: {newquestion3}
Possible Answers: {choices1}

Remember, you must provide only items A-C in the correct format. Do not repeat yourself or elements of the prompt. Do not give extraneous information other than items A-C.}

\clearpage

\newpage
\section{Robustness Checks}

\subsection{Unpaired Benchmark Results}
\label{app:unpaired}
For completeness, we provide unpaired benchmark results, mirroring the paired table in the paper. As can be seen, there is moderate sensitivity to language perturbation and ordering, with unpaired results besting paired results by 10-12\%, in general. Again, Gemini 2.5 Flash dominates the results, with PerceptionLM and InternVL2.5 ahead of GPT-4o on our reasoning subsection.

\begin{table}[!htb]
    \setlength{\tabcolsep}{2pt}
    \renewcommand{\arraystretch}{1}
    
    \small
    \begin{tabular}{ll*{8}c}
        \toprule
        \multicolumn{3}{c}{} & \multicolumn{2}{c}{\textbf{Large/Closed}} & \multicolumn{4}{c}{\textbf{7-8b/Open}} & \\
        \cmidrule(lr){4-5} \cmidrule(lr){6-9}
        Category & 
        Difficulty & 
        N &
        \rotatebox{45}{GPT-4o} & 
        \rotatebox{45}{Gemini 2.5 Flash} & 
        \rotatebox{45}{InternVL2.5} & 
        \rotatebox{45}{LLaVa-OneVision} & 
        \rotatebox{45}{Perception-LM} & 
        \rotatebox{45}{Qwen2.5VL} & 
        \rotatebox{45}{Human} \\
        \midrule
        Anticipation & Easy & 220 & 61.82 & \textbf{64.55} & 59.09 & 50.91 & 60.91 & 57.27 & \textit{94.26} \\
        & Med & 128 & 42.19 & 46.88 & 46.09 & 35.94 & \textbf{56.25} & 41.41 & \textit{77.52} \\
        & Hard & 68 & 25.00 & \textbf{52.94} & 35.29 & 44.12 & 36.76 & 39.71 & \textit{60.17} \\
        \midrule
        Counterfactual & Easy & 56 & 73.21 & \textbf{91.07} & 57.14 & 69.64 & 69.64 & 62.50 & \textit{92.76} \\
        & Med & 68 & \textbf{57.35} & 50.00 & 47.06 & 50.00 & 54.41 & 50.00 & \textit{77.17} \\
        & Hard & 54 & 42.59 & 46.30 & 55.56 & \textbf{57.41} & 51.85 & 55.56 & \textit{58.96} \\
        \midrule
        Descriptive & Easy & 452 & 79.42 & \textbf{88.50} & 63.27 & 70.13 & 64.38 & 74.56 & \textit{95.96} \\
        & Med & 88 & 67.05 & \textbf{75.00} & 54.55 & 52.27 & 56.82 & 61.36 & \textit{76.19} \\
        & Hard & 36 & 38.89 & 52.78 & 36.11 & \textbf{55.56} & 41.67 & 52.78 & \textit{59.02} \\
        \midrule
        Hypothetical & Easy & 30 & \textbf{80.00} & 76.67 & 70.00 & 73.33 & 70.00 & 60.00 & \textit{92.63} \\
        & Med & 62 & 50.00 & \textbf{67.74} & 38.71 & 37.10 & 43.55 & 54.84 & \textit{77.32} \\
        & Hard & 42 & 30.95 & \textbf{57.14} & 52.38 & 30.95 & 52.38 & 38.10 & \textit{61.94} \\
        \midrule
        Planning & Easy & 150 & 73.33 & \textbf{78.00} & 76.67 & 65.33 & 77.33 & 63.33 & \textit{94.76} \\
        & Med & 84 & 55.95 & \textbf{79.76} & 65.48 & 67.86 & 71.43 & 58.33 & \textit{78.46} \\
        & Hard & 48 & 52.08 & \textbf{62.50} & 39.58 & 50.00 & 47.92 & 43.75 & \textit{60.27} \\
        \midrule
        \multicolumn{10}{c}{Aggregates} \\
        \midrule
        Anticipation & -- & 416 & 49.76 & \textbf{57.21} & 51.20 & 45.19 & 55.53 & 49.52 & \textit{83.54} \\
        Counterfactual & -- & 178 & 57.88 & \textbf{61.82} & 52.82 & 58.44 & 58.44 & 55.63 & \textit{76.57} \\
        Descriptive & -- & 576 & 75.00 & \textbf{84.20} & 60.24 & 66.49 & 61.80 & 71.18 & \textit{90.63} \\
        Hypothetical & -- & 134 & 50.74 & \textbf{66.41} & 49.99 & 43.28 & 52.23 & 50.74 & \textit{75.92} \\
        Planning & -- & 282 & 64.50 & \textbf{75.85} & 66.99 & 63.44 & 70.53 & 58.48 & \textit{83.99} \\
        \midrule
        Reasoning & -- & 1010 & 55.44 & \textbf{64.45} & 55.74 & 52.37 & 59.79 & 53.26 & \textit{81.43} \\
        All & -- & 1586 & 62.55 & \textbf{71.63} & 57.38 & 57.50 & 60.53 & 59.77 & \textit{84.78} \\
        \bottomrule
    \end{tabular}
    \caption{Unpaired performance metrics by model across benchmark question categories and difficulties. Reasoning questions refer to all question types except for \textit{descriptive}.}
    \label{tab:single-performance}
\end{table}

\begin{table}[!h]
    \setlength{\tabcolsep}{4pt}
    \renewcommand{\arraystretch}{1.3}
        \caption{Paired performance metrics by model across benchmark question categories and difficulties. \textit{Blind} means that no video was included in the input, while \textit{Single Frame} includes only the last frame.}
    \small
    \begin{tabular}{llc *{3}{c}}
        \toprule
         & 
         & &
        \textbf{Blind} & 
        \multicolumn{2}{c}{\textbf{Single Frame}} \\
        \cmidrule(lr{-2pt}){4-4}\cmidrule(lr{2pt}){5-6}
        Category & Difficulty & N &
        \makebox[2.2cm][c]{\rotatebox{45}{Qwen2.5-72B-Instruct}} & 
        \makebox[2.2cm][c]{\rotatebox{45}{Qwen2.5VL-7B-Instruct}} & 
        \makebox[2.2cm][c]{\rotatebox{45}{Human}} \\
        \midrule
        Anticipation & Easy & 110 & 14.55 & 10.91 & \textit{76.53} \\
        & Med & 64 & 23.44 & 28.13 & \textit{69.19} \\
        & Hard & 34 & 23.53 & 14.71 & \textit{59.92} \\
        \midrule
        Counterfactual & Easy & 28 & 53.57 & 42.86 & \textit{77.42} \\
        & Med & 34 & 23.53 & 17.65 & \textit{67.78} \\
        & Hard & 27 & 29.63 & 25.93 & \textit{62.45} \\
        \midrule
        Descriptive & Easy & 226 & 23.45 & 19.91 & \textit{75.83} \\
        & Med & 44 & 31.82 & 29.55 & \textit{66.09} \\
        & Hard & 18 & 27.78 & 22.22 & \textit{67.73} \\
        \midrule
        Hypothetical & Easy & 15 & 40.00 & 46.67 & \textit{82.32} \\
        & Med & 31 & 22.58 & 22.58 & \textit{71.50} \\
        & Hard & 21 & 19.05 & 19.05 & \textit{63.22} \\
        \midrule
        Planning & Easy & 75 & 32.00 & 22.67 & \textit{76.92} \\
        & Med & 42 & 23.81 & 23.81 & \textit{65.95} \\
        & Hard & 24 & 25.00 & 20.83 & \textit{59.88} \\
        \midrule
        \multicolumn{5}{c}{Aggregates} \\
        \midrule
        Anticipation & -- & 208 & 18.75 & 16.83 & \textit{71.56} \\
        Counterfactual & -- & 89 & 34.83 & 28.09 & \textit{69.19} \\
        Descriptive & -- & 288 & 25.00 & 21.53 & \textit{73.84} \\
        Hypothetical & -- & 67 & 25.37 & 26.87 & \textit{71.33} \\
        Planning & -- & 141 & 28.37 & 22.70 & \textit{70.75} \\
        \midrule
        Reasoning & -- & 505 & 25.15 & 21.78 & \textit{70.89} \\
        All & -- & 793 & 25.09 & 21.69 & \textit{71.83} \\
        \bottomrule
    \end{tabular}
    \label{tab:ablation-performance}
\end{table}

\subsection{Blind and Single Frame Results}
\label{app:blind}
For robustness, we conducted tests of a \textbf{Blind LLM} to see the susceptibility of the benchmark to the baseline knowledge present in a language backbone. We selected Qwen2.5-72B-Instruct as it is similar in size to Llama 3.1-70B-Instruct, the model we used in construction, while remaining different enough to provide a check on our efforts to remove linguistic shortcuts. Results can be seen in the table below. 

While we made every effort to remove linguistic clues that would allow an LLM, alone, to perform well, material information remains. We believe a couple factors explain this: first, our LLM was different -- Llama is trained and optimized differently than Qwen -- so we believe some residual information is the result of scrubbing the dataset with a model that has different (and somewhat non-overlapping) sensitivities; second, we do our LLM filtering after our initial question creation and pipeline -- we subsequently perform a quality assurance round with human annotators. We do not perform another blind LLM filtering step, so our human annotators potentially re-introduce linguistic hints. Still, average performance on the blind test is $\approx25\%$, whether we are talking about reasoning tasks or the full benchmark -- this is significantly worse than any of the video models tested, indicating most items in the benchmark still rely on visual information to be answered correctly. 

We also test whether or not the benchmark requires material video understanding, or if an image is sufficient. \textbf{Single Frame} gives the final frame of the question clip as input. We test Qwen2.5VL (the same 7b model from the paper) and humans to provide bookends on the susceptibility of the benchmark to this shortcut. Human testing was similar to our other human baseline (though with only $87$ annotators). Again, since no annotator took the full benchmark, we give bootstrap estimates (10,000 draws stratified by question category and difficulty). 

In general, we see that both ablations (blind with a strong language backbone or a single frame with a more modest one) perform similarly poorly ($21$-$25\%$), with the stronger language backbone adding slightly more to performance (we suspect that the benchmark does require meaningful commonsense world knowledge better captured in the larger blind Qwen2.5 model). Humans perform well even if only given a single frame, which may be difficult to avoid when designing questions about real-world physical activity.

\newpage

\section{Question Pairing}
\label{app:pairing}
We introduced question paring for two main purposes: First, as described in the paper body, perturbing and reordering answer sets in paired questions allowed us to directly test for sensitivity to minor language differences and order. Second, our question items required considerable cost and effort to generate, limiting the benchmark's length. Pairing questions is an inexpensive and viable strategy to decrease variance and improve our ability to register small improvements in model performance, as outlined below.

Variance for a model's run on a benchmark can be be thought of as the contribution from the questions where the model knows the answer (we'll call this fraction of the benchmark \begin{math}k\end{math}) and the unknown portion of the benchmark (we'll call \begin{math}u = (1-k)\end{math}). Both of these quantities are unobserved.

Recall:

1. binomial variance \begin{math}\sigma^2\end{math} is a function of \begin{math}p\end{math} (probability correct)

\begin{math}\sigma^2 = p(1-p)\end{math}

and, 

2. overall variance from two sources is:

\begin{math}\sigma^2 = w_1^2\sigma_1^2 + w_2^2\sigma_2^2\ + 2w_1w_2cov(\sigma_1,\sigma_2)\end{math}

substituting here:

\begin{math}w_1 = k\end{math} the known fraction

\begin{math}w_2 = u\end{math} the unknown fraction

\begin{math}\sigma_1^2 = \sigma_k^2 = p_k(1-p_k) = 100\%(1-100\%) = 0 \end{math}

\begin{math}\sigma_2^2 = \sigma_{u}^2 p_{u}(1-p_{u})\end{math} where \begin{math}p_{u}\end{math} is the chance/guess rate

\begin{math}cov(\sigma_1,\sigma_2) = 0\end{math} since \begin{math}\sigma_1 = 0 \end{math} and \begin{math}\sigma_2 \end{math} is noise

so the variance is entirely dependent on the chance/guess probability for the unknown portion of the benchmark and simplifies to

\begin{math}\sigma^2 = w_2^2\sigma_2^2 = u^2p_{u}(1-p_{u})\end{math}

\vspace{5mm}

From the statistical testing literature:

\begin{math} d = \sigma Z \sqrt{2/n} \end{math} 

where:

\begin{math}d\end{math} is the detectable effect size,

\begin{math}n\end{math} is benchmark size, 

\begin{math}\sigma\end{math} is standard deviation (in our case, we know this is \begin{math}\sigma_u\end{math} since the unknown portion contributes all the variance), 

\begin{math}Z\end{math} is the combined Z-scores for fixed protection from type I and type II errors. 

It is easy to show \begin{math}d\end{math} is reduced by pairing, as it is entirely dependent on the ratio of variances between an unpaired and paired benchmark.

For our five option multiple choice questions:

\begin{math}p_{paired} = 0.2^2 = 0.04 \rightarrow \sigma_{paired} = \sqrt{.04(1-.04)} = 0.196\end{math}, 

vs.

\begin{math}p_{unpaired} = 0.2 \rightarrow \sigma_{unpaired} = \sqrt{.2(1-.2)} = 0.4 \end{math} 

so \begin{math}d_{paired}\end{math} is approximately \begin{math}\frac{d_{unpaired}}{2}\end{math} or the equivalent of a test approximately four times larger (rather than being just twice as large from pairing)
\end{document}